\documentclass{article}

\usepackage{microtype}
\usepackage{graphicx}
\usepackage{subcaption}
\usepackage{booktabs}%

\usepackage{hyperref}
\usepackage{caption}
\PassOptionsToPackage{numbers}{natbib}
\usepackage[main,preprint]{neurips_2026}
\usepackage[utf8]{inputenc}%
\usepackage[T1]{fontenc}%

\usepackage{url}%

\usepackage{amsfonts}%
\usepackage{nicefrac}%
\usepackage{microtype}%
\usepackage{xcolor}
\usepackage{float}

\usepackage{amsmath}
\usepackage{amssymb}
\usepackage{mathtools}
\usepackage{amsthm}

\usepackage[capitalize,noabbrev]{cleveref}
\usepackage{adjustbox}
\usepackage{multirow}
\usepackage{booktabs}
\usepackage{tabularx}%
\usepackage{xspace}
\usepackage{tcolorbox}
\newcommand{\apples}{RL4HS\xspace}

\newcommand{\myquote}[1]{``#1''} 
\theoremstyle{plain}

\theoremstyle{definition}

\theoremstyle{remark}

\title{SOPO: Socratic Guided Policy Optimization for Span-Level Hallucination Detection}

\author{
	\textbf{Jiawei Dong}\textsuperscript{1} \quad
	\textbf{Zilong Bai}\textsuperscript{1} \quad
	\textbf{Pengtian Zhu}\textsuperscript{1} \quad 
	\textbf{Peng Zhou}\textsuperscript{1} \quad \\
	\textsuperscript{1}Patsnap Pte. Ltd., Suzhou, China \quad \\
	\texttt{\{dongjiawei,baizilong,zhupengtian,tonyzhoupeng\}@patsnap.com}
}

\begin{document}

\maketitle

\begin{abstract}

Large Language Models (LLMs) have gained widespread adoption in various natural language processing tasks, but suffer from hallucination issues where they generate unfaithful or inconsistent content. While Reinforcement Learning with Verifiable Rewards (RLVR) has shown promising applications in reasoning tasks, its application to span-level hallucination detection is challenged by localization ambiguity, reward sparsity on hard examples, and uncontrolled length inflation. To address these issues, we introduce Socratic Guided Policy Optimization (SOPO), a multi-turn RL framework for robust hallucination detection. The key innovation of SOPO is a Socratic feedback mechanism, where a teacher model guides the reasoning trajectories during rollout by providing indirect hints inferred from the ground truth. This teacher-guided approach enables the model to generate more effective reasoning trajectories for hard examples, thus alleviating reward sparsity. Furthermore, we incorporate Turn-based Advantage Scaling to penalize verbosity bias, encouraging the model to develop efficient and autonomous reasoning capabilities. Empirical results on RAGTruth demonstrate that SOPO (4B) achieves state-of-the-art performance, surpassing leading proprietary models (e.g., GPT-5) and standard RL baselines, while obtaining higher reported mean scores than GRPO on four out-of-domain benchmarks, with mixed evidence against CAPO. We release our training and evaluation code alongside the 1.7B and 4B SOPO models at \url{https://github.com/patsnap/SOPO}.
\end{abstract}

\section{Introduction}
\label{intro}
Retrieval-Augmented Generation (RAG) has emerged as a dominant paradigm for mitigating hallucinations in Large Language Models (LLMs) by grounding generation in external knowledge. However, even with retrieved context, LLMs inevitably produce factually inconsistent content \cite{sun2025redeepdetectinghallucinationretrievalaugmented, zhou2024metacognitiveretrievalaugmentedlargelanguage}. Detecting these hallucinations at a fine-grained level, specifically by pinpointing the exact hallucinated spans, is critical for ensuring the reliability and safety of real-world applications.

While Supervised Fine-Tuning (SFT) has established a baseline for this task, recent research has shifted towards Reinforcement Learning (RL), particularly outcome-based methods like Reinforcement Learning with Verifiable Rewards (RLVR), to further align models with complex reasoning requirements. 

Prior research has largely focused on case-level (binary) hallucination detection using SFT or RL \cite{eliav2025clattercomprehensiveentailmentreasoning}. For instance, FaithLens \cite{si2026faithlensdetectingexplainingfaithfulness} demonstrated that an 8B model, trained via a rigorous four-stage pipeline, could rival proprietary models on case-level tasks. To address the granularity gap, RL4HS \cite{su2025learningreasonhallucinationspan} pioneered the application of RL to span-level detection, utilizing Class-Aware Policy Optimization (CAPO) to outperform proprietary models.

While recent guided-RL and iterative-refinement paradigms aim to enhance reasoning, they exhibit structural mismatches for objective, fine-grained hallucination detection. Specifically, unified models \cite{xu2026stepwisethinkcritiqueunifiedframework} compromise diagnostic objectivity due to role conflicts when evaluating their own generations; Existing Socratic-RL frameworks \cite{wu2025socraticrlnovelframework} rely on post-hoc distillation instead of providing dynamic, instance-specific hints during active RL exploration; and direct feedback methods \cite{zhang2026critiquegrpoadvancingllmreasoning} supply explicit single-turn corrections, causing ground-truth leakage that bypasses autonomous reasoning process.

Consequently, in the absence of suitable guided frameworks, standard RLVR continues to serve as the default baseline. Yet, our analysis demonstrates that applying such scalar-reward RL directly to fine-grained span detection is inherently hindered by three fundamental limitations:

\paragraph{Ambiguity of Localization Signals.} From a reward modeling perspective, systematic analyses of RLHF have highlighted the fundamental limitations of scalar reward models regarding expressiveness and feedback sparsity \cite{chaudhari2024rlhfdecipheredcriticalanalysis}. In the specific context of span-level hallucination detection, this limitation is amplified: a single scalar reward is forced to represent a complex, comprehensive evaluation of accuracy across three distinct dimensions, namely boundary alignment, semantic completeness, and factual discrimination. Consequently, using such a coarse-grained proxy for a fine-grained task introduces significant ambiguity. For instance, particularly in samples where the target span consists of only a few words, a deviation of a single token can trigger drastic drops in the reward score (e.g., F1 dropping from 1.0 to 0.2), resulting in instability in the training process. This instability is supported in RL4HS \cite{su2025learningreasonhallucinationspan}, where the direct application of span-F1 as a reward required the introduction of CAPO to mitigate reward imbalance. This evidence further confirms that coarse scalar signals induce optimization instability and bias in hallucination span detection.

\paragraph{Reward Sparsity on Hard Examples.} The model struggles to learn from hard examples due to the sparsity of the reward signal. With a small group size ($G$), the model consistently receives a zero reward across epochs for difficult queries, causing performance plateauing in the later stages of training. Conversely, simply increasing $G$ to improve exploration introduces a trade-off: it generates a large number of \myquote{zero-advantage} groups for easy samples (where all rollouts are correct), which provides negligible gradient information \cite{zhang2025edgegrpoentropydrivengrpoguided, nan2025ngrponegativeenhancedgrouprelative}. This results in wasted computational resources and significantly reduces overall training efficiency \cite{feng2025dontwastemistakesleveraging}.

\paragraph{Uncontrollable Reasoning Length Inflation.} During RLVR training, length of reasoning chains exhibits a tendency to inflate uncontrollably as training progresses. This phenomenon is not merely accidental but appears to be a structural limitation of current RL objectives. Prior analyses suggest that this verbosity is largely an optimization artifact: PPO/GRPO-style objectives systematically drive the policy toward longer outputs, particularly when the model struggles with correctness, leading to a persistent preference for lengthy generations \cite{fatemi2025concisereasoningreinforcementlearning, dai2025stablereinforcementlearningefficient}. This aligns with recent findings on \myquote{systematic length inflation,} where models trade off efficiency for length by generating redundant \myquote{filler} tokens without improving accuracy \cite{shrivastava2025samplethinklessgroup}. Consequently, without explicit regularization, this length inflation is bounded only by the context window. For practical applications, such verbosity creates severe computational bottlenecks and drastically increases inference latency.

To address these limitations, drawing inspiration from human pedagogy, we propose \textbf{Socratic Guided Policy Optimization (SOPO)}. Central to SOPO is a teacher LLM that engages the student in a multi-turn Socratic dialogue, offering step-wise analysis and guiding questions based on the student's incorrect reasoning steps \textit{without revealing the ground truth}. This iterative interaction transforms sparse outcome rewards into dense, constructive supervision. Complementing this, we introduce a Turn-based Advantage Scaling mechanism that explicitly penalizes verbosity and reliance on multi-turn feedback, encouraging the model to achieve efficient, autonomous detection in the initial turn.

We validate the effectiveness of our approach on the RAGTruth benchmark \cite{niu2024ragtruthhallucinationcorpusdeveloping}. Empirical evaluations demonstrate that SOPO establishes a new state-of-the-art with remarkable training efficiency. 

Our main contributions are summarized as follows:
\begin{itemize}
	
	\item We identify the limitations of vanilla RLVR in span-level hallucination detection task and propose SOPO, a framework that incorporates a Socratic-style feedback mechanism. By offering analysis and guiding questions without ground truth leakage, we transform the training process from passive reward reception to active iterative reasoning refinement, effectively preventing shortcut learning while encouraging intrinsic self-correction.
	
	\item We introduce Turn-based Advantage Scaling, a simple yet effective technique that explicitly penalizes verbosity and reliance on multi-turn feedback. This mechanism reshapes the advantage to encourage autonomous and efficient reasoning, addressing the issue of reasoning length inflation common in RL training.
	
	\item Experiments on the RAGTruth and other out-of-distribution benchmark demonstrate that SOPO (4B) achieves superior performance. Notably, despite its compact size (4B parameters), SOPO outperforms larger proprietary models (e.g., GPT-5, o3) and surpasses the previous state-of-the-art by 4.4 points in overall F1 on RAGTruth test set, implying the effectiveness of our Socratic RL paradigm.
	
\end{itemize}

\section{Span-level Hallucination Detection}
\subsection{Task}

We focus on the task of {Span-level Hallucination Detection} within Retrieval-Augmented Generation (RAG) scenarios, encompassing tasks such as summarization and knowledge-intensive question answering. 
Formally, let $\mathcal{C}$ denote the retrieved context document and $\mathcal{Y} = (y_1, y_2, \dots, y_T)$ denote the model-generated response consisting of $T$ tokens or characters. The objective is to identify a set of unreliable spans $\hat{\mathcal{S}} = \{s_1, s_2, \dots, s_n\}$, where each span $s_k$ represents a sequence within $\mathcal{Y}$ that contradicts or is unsubstantiated by the context $\mathcal{C}$. Each span is defined by its start and end character indices, i.e., $s_k = [start_k, end_k)$. If the response contains no hallucinations, $\hat{\mathcal{S}}$ should be an empty set.

\subsection{Metric}
To rigorously evaluate the localization accuracy of the detected hallucinations, we adopt the {Span-level F1 score}, a standard metric for span-level hallucination detection task \cite{niu2024ragtruthhallucinationcorpusdeveloping}.

Let $\mathcal{S}^*$ represent the set of ground-truth hallucination character indices, and $\hat{\mathcal{S}}$ represent the set of predicted character indices. We treat the spans as sets of integers representing character positions. The Precision ($P$) and Recall ($R$) are calculated based on the character-level overlap between the prediction and the ground truth, where $|\cdot|$ denotes the cardinality of the set (i.e., the total number of characters), and $\cap$ denotes the intersection operation. The final metric is given by:
\begin{equation}
	P = \frac{|\hat{\mathcal{S}} \cap \mathcal{S}^*|}{|\hat{\mathcal{S}}|}, \quad 
	R = \frac{|\hat{\mathcal{S}} \cap \mathcal{S}^*|}{|\mathcal{S}^*|}, \quad 
	\text{span-F1} = \frac{2 \cdot P \cdot R}{P + R}
\end{equation}

This metric encourages the model to cover the hallucinated content comprehensively (high recall) while minimizing the inclusion of factual text (high precision).

\section{SOPO: Socratic Guided Policy Optimization}
\label{sec:sopo}
\begin{figure*}[h!]
	\centering
	\hfill
	\vspace{0pt}%
	\includegraphics[width=\linewidth]{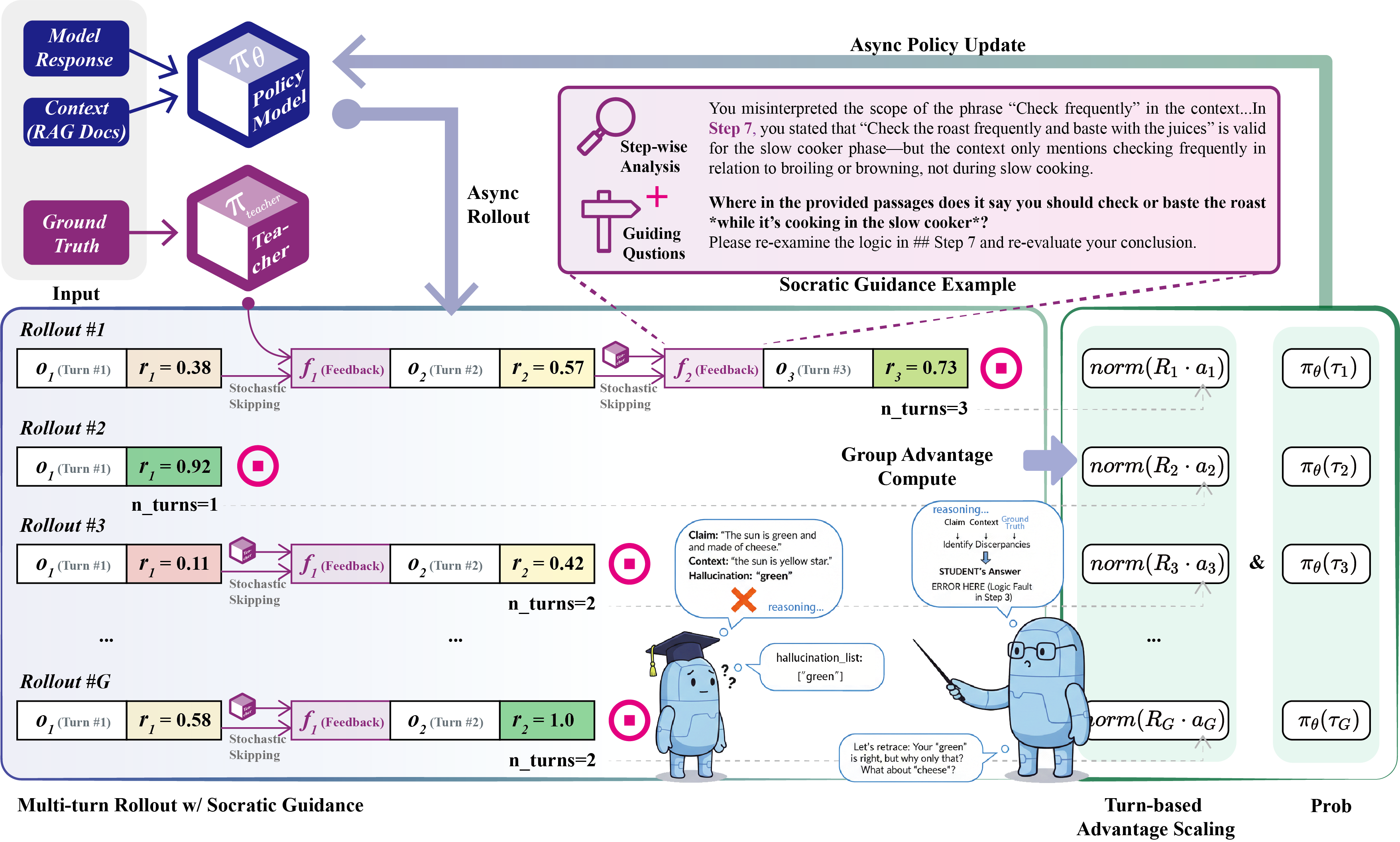}
	\caption{Overview of SOPO.}
	\label{fig:workflow}%
\end{figure*}

To address the aforementioned limitations of RLVR on hallucination detection, we introduce a heuristic multi-turn training strategy named \textbf{Socratic Guided Policy Optimization (SOPO)}. By incorporating a Teacher LLM, SOPO shifts the paradigm from scalar reward maximization to iterative refinement. At each turn, the teacher evaluates the student's reasoning and generates a detailed analysis of specific incorrect steps, and then raises a guiding question to encourage the student to self-correct. Through this multi-turn, Socratic-style feedback, we guide the model toward dense, step-wise optimization, thereby alleviating the performance plateauing often observed under sparse reward settings. The overall workflow is shown in Figure~\ref{fig:workflow}.

\subsection{Formalization of Socratic Interaction}

We formulate the hallucination detection task as a multi-turn decision-making process involving a student policy $\pi_\theta$ and a frozen teacher policy $\pi_{teacher}$. Let $x$ denote the input context (including source documents and the response to be verified) and $y^*$ represent the ground-truth hallucination spans.

The turns of interaction is limited to $T$. At each turn $t$ ($1 \le t \le T$), the process consists of three phases: generation, evaluation, and feedback.

\paragraph{Student Generation.}
Given the input $x$ and the interaction history $H_{t-1}$, the student model generates a reasoning path and a prediction $o_t$:
\begin{equation}
	o_t \sim \pi_\theta(\cdot \mid x, H_{t-1})
\end{equation}
where $H_0 = \emptyset$, and $H_{t-1}$ accumulates the trajectory of previous outputs and feedback.

\paragraph{Evaluation.}
At each turn, we evaluate the quality of $o_t$ against the ground truth $y^*$ using the span-level F1 reward function $R(\cdot)$. The interaction process terminates early if the reward exceeds a predefined threshold $\delta$ (e.g., $\delta=0.6$), serving as a regularization measure to prevent overfitting to potential annotation noise in the training set. Furthermore, to prevent the student policy from developing a rigid dependency on teacher's feedback, we stochastically provide feedback: for lower-scoring responses ($R(o_t, y^*) < \delta$), the teacher provides feedback with probability $p_{\text{feedback}}$; otherwise, the feedback phase is skipped.

\paragraph{Socratic Feedback Generation.}
The teacher model $\pi_{teacher}$ acts as a Socratic guide, unlike standard correction methods that directly provide $y^*$, the teacher generates a feedback signal $f_t$ conditioned on the student's incorrect reasoning, specifically designed to be \textit{answer-agnostic}, encouraging the student to derive the conclusion independently. The feedback consists of two components: a step-wise analysis ($\mathcal{A}$) and a guiding question ($\mathcal{G}$):
\begin{equation}
	f_t = [\mathcal{A}_t; \mathcal{G}_t] \sim \pi_{teacher}(\cdot \mid x, o_t, \mathcal{P}_{\text{Socratic}})
\end{equation}
where $\mathcal{P}_{\text{Socratic}}$ represents the system prompt enforcing the constraints against leaking the ground truth.

The history is then updated to $H_t = H_{t-1} \cup \{o_t, f_t\}$, and the student attempts to refine its reasoning in the next turn $t+1$. The complete trajectory $\tau$ for a sample is thus defined as:
\begin{equation}
	\tau = \{ (o_1, r_1), (f_1), (o_2, r_2), (f_2), \dots, (f_{k-1}), (o_k, r_k) \}
\end{equation}
where $r_t$ is the reward score of $o_t$ and $k \le T$ is the turn where the model either succeeds or reaches the turn limit. Our optimization goal is to maximize the reward of these trajectories, weighted by our turn-based advantage scaling mechanism.

\subsection{GRPO with Turn-based Advantage Scaling}

We adopt Group Relative Policy Optimization \cite{shao2024deepseekmathpushinglimitsmathematical} as our reinforcement learning backbone. Unlike PPO, GRPO eliminates the explicit value function critic and estimates the baseline directly from group scores. The learning objective is defined as:

\begin{equation}
	\mathcal{L}_{\text{GRPO}}(\theta) = \mathbb{E}_{\tau \sim \pi_\theta} \bigg[ \min \Big( r_\theta(\tau) A(\tau), \text{clip}\left(r_\theta(\tau), 1 - \varepsilon, 1 + \varepsilon\right) A(\tau) \Big) \bigg]
\end{equation}
where $\tau$ denotes a trajectory sampled from $\pi_\theta$, and $r_\theta(\tau)$ is the probability ratio between the updated and reference policies.

In original GRPO, the advantage $A_i$ for the $i$-th output in a group of size $G$ is typically computed by directly normalizing the raw rewards $\mathbf{R} = \{R_1, \dots, R_G\}$:
\begin{equation}
	A_i = \frac{R_i - \text{mean}(\mathbf{R})}{\text{std}(\mathbf{R}) + \epsilon}
\end{equation}
However, relying solely on unpenalized task rewards fails to distinguish and penalize trajectories that achieve identical scores but consume significantly more interactions. This flaw inadvertently encourages interaction redundancy, allowing the model to secure high rewards by heavily relying on extended dialogue rather than solving the task autonomously.

To counteract this, we introduce a {Turn-based Advantage Scaling} mechanism. Given a sampled trajectory $\tau_i$ consisting of a total of $k_i$ turns ($1 \le k_i \le T$), we define a penalty hyperparameter $p_{\text{penalty}}$ to compute a scaling factor $\alpha_i$, which modifies the advantage as follows:

\begin{equation}
	\begin{gathered}
		\alpha_i = \text{clip}\left(1 - (k_i - 1) \cdot p_{\text{penalty}}, 0, 1\right), \quad R_{\text{scaled}, i} = \alpha_i \cdot R_i, \\
		A_i = \frac{R_{\text{scaled}, i} - \text{mean}(\mathbf{R}_{\text{scaled}})}{\text{std}(\mathbf{R}_{\text{scaled}}) + \epsilon}
	\end{gathered}
\end{equation}

where $\mathbf{R}_{\text{scaled}} = \{R_{\text{scaled}, 1}, \dots, R_{\text{scaled}, G}\}$ represents the set of scaled rewards within the group. This pre-normalization intervention effectively incorporates our heuristic constraints without violating the mathematical integrity of the zero-mean advantage estimator (see Appendix ~\ref{sec:adv_scale}). With these modifications, we explicitly penalize reliance on multi-turn feedback and encourage autonomous reasoning without altering the fundamental definition of the task metric.

\subsection{Span-Level F1 Reward}

To align with our fine-grained objective, we formulate the reward directly based on the Span-level F1 score. Let $\hat{\mathcal{S}}$ denote the set of predicted hallucination spans and $\mathcal{S}^*$ represent the ground-truth spans. The reward at turn $t$ is:
\begin{equation}
	r_t = 
	\begin{cases} 
		1, & \text{if } \hat{\mathcal{S}} = \emptyset \text{ and } \mathcal{S}^* = \emptyset, \\
		\text{span-level F1}(\hat{\mathcal{S}}, \mathcal{S}^*), & \text{otherwise}.
	\end{cases}
\end{equation}

In multi-turn training, we employ a turn-averaged reward to encourage early error correction:
\begin{equation}
	R_i = \frac{1}{k_i} \sum_{t=1}^{k_i} r_t
\end{equation}
where $k_i$ is the total number of turns in trajectory $\tau_i$. Complementary to Advantage Scaling, this acts as a local quality constraint to prevent the model from intentionally failing early turns to solicit hints (detailed in Appendix~\ref{sec:dual_penalty_synergy}).

\section{Experimental Setup}
\label{sec:exp_setup}
\paragraph{Dataset.}To directly validate the effectiveness of our proposed method in Retrieval-Augmented Generation (RAG) scenarios, we conduct experiments on the RAGTruth \cite{niu2024ragtruthhallucinationcorpusdeveloping} benchmark. This dataset encompasses three distinct tasks: Summarization, Question Answering (QA), and Data-to-Text. The training set contains 15,090 samples, and the test set contains 2,700 samples. Each sample in the dataset consists of paired source documents, LLM-generated response, and human-annotated hallucination spans.

\paragraph{Model.}In our experiments, we use the Qwen3-4B-2507 reasoning model \cite{yang2025qwen3technicalreport} as the policy model and the Qwen3-Next-80B-A3B-Instruct as the feedback model. The parameters of the feedback model remain fixed throughout all experiments.

\paragraph{Cold-start.}To accelerate convergence and address the cold-start problem, we randomly split the RAGTruth training set into a 5,090-sample subset for SFT and a 10,000-sample subset for RL training. We employ Gemini-2.5-Flash to generate multiple responses for each SFT prompt. Then, we apply rejection sampling to select responses with the highest Span-level F1 scores against the ground truth to construct the cold-start SFT dataset. Both SOPO and all baseline models are initialized from this SFT checkpoint for the subsequent RL training.

\begin{table*}[ht!]
	\centering
	
	\caption{\textbf{Span-level hallucination detection results on RAGTruth.} 
		We compare our results with several approaches presented in Multi-View \cite{ogasa2025hallucinatedspandetectionmultiview}, RL4HS \cite{su2025learningreasonhallucinationspan}, LettuceDetect \cite{kovács2025lettucedetecthallucinationdetectionframework}. Best scores are in bold, second best scores are underlined.  
	}
	\begin{adjustbox}{width=\linewidth}
		
		\begin{tabular}{c|ccc|ccc|ccc|ccc}
			\toprule
			\multicolumn{1}{c|}{\multirow{2}{*}{\textbf{Model}}} & \multicolumn{3}{c|}{\textbf{Summarization}} & \multicolumn{3}{c|}{\textbf{Question Answering}} & \multicolumn{3}{c|}{\textbf{Data-to-Text}} & \multicolumn{3}{c}{\textbf{Overall}} \\
			\multicolumn{1}{c|}{} & F1 & Precision & Recall & F1 & Precision & Recall & F1 & Precision & Recall & F1 & Precision & Recall \\ \midrule
			\multicolumn{13}{l}{\textit{\textbf{Proprietary models}}} \\ \midrule
			\multicolumn{1}{c|}{GPT-4o-mini w/ CoT} & 38.4 & 43.4 & 34.4 & 27.3 & 33.7 & 23.0 & 33.7 & 34.2 & 33.2 & 32.4 & 35.9 & 29.5 \\
			\multicolumn{1}{c|}{GPT-5-mini w/ CoT} & 43.9 & 33.0 & \underline{65.5} & 47.2 & 37.9 & 62.7 & 42.5 & 29.8 & \textbf{74.7} & 44.2 & 32.8 & \underline{67.7} \\
			\multicolumn{1}{c|}{GPT-5 w/ CoT} & 36.5 & 24.9 & \textbf{68.4} & 44.4 & 32.1 & \underline{71.8} & 45.7 & 33.2 & \underline{73.5} & 42.9 & 30.7 & \textbf{71.1} \\
			\multicolumn{1}{c|}{o3 w/ CoT} & 48.5 & 40.7 & 60.1 & 49.9 & 43.4 & 58.9 & 55.2 & 45.6 & 70.0 & 51.8 & 43.8 & 63.2 \\ 
			\midrule
			\multicolumn{13}{l}{\textit{\textbf{Non-Reasoning models}}} \\ 
			
			\midrule
			\multicolumn{1}{c|}{Qwen3-4B-Instruct-2507} & 43.4 & 46.6 & 40.7 & 33.7 & 27.4 & 43.8 & 35.7 & 28.3 & 48.3 & 36.2 & 30.2 & 45.0 \\
			\multicolumn{1}{c|}{Qwen3-30B-A3B-Instruct-2507} & 35.1 & 37.5 & 33.0 & 19.0 & 20.4 & 17.8 & 28.5 & 29.1 & 28.0 & 26.5 & 27.7 & 25.3 \\
			\multicolumn{1}{c|}{Qwen3-32B-Instruct} & 44.0 & 59.7 & 34.9 & 19.4 & 33.3 & 13.7 & 34.4 & 44.3 & 28.2 & 31.5 & 44.7 & 24.3 \\
			\multicolumn{1}{c|}{Qwen3-Next-80B-A3B-Instruct} & 48.2 & 45.7 & 51.1 & 33.8 & 26.6 & 46.3 & 42.5 & 39.5 & 45.9 & 39.9 & 34.5 & 47.2 \\
			
			\midrule
			\multicolumn{13}{l}{\textit{\textbf{Reasoning models}}} \\ \midrule
			
			\multicolumn{1}{c|}{Qwen3-8B} & 34.7 & 42.2 & 29.5 & 30.5 & 32.0 & 29.1 & 20.3 & 45.2 & 13.1 & 27.8 & 37.2 & 22.1 \\
			\multicolumn{1}{c|}{Qwen3-14B} & 35.8 & 36.9 & 34.9 & 30.6 & 30.7 & 30.6 & 34.8 & 40.9 & 30.4 & 33.3 & 35.8 & 31.1 \\ 
			\multicolumn{1}{c|}{Qwen3-4B-2507} & 47.0 & 44.0 & 50.4 & 34.7 & 33.3 & 36.1 & 49.8 & 48.0 & 51.7 & 43.6 & 41.8 & 45.7 \\
			\multicolumn{1}{c|}{Qwen3-30B-A3B-2507} & 46.9 & 43.8 & 50.6 & 37.2 & 33.7 & 41.5 & 51.0 & 51.7 & 50.3 & 44.8 & 42.6 & 47.1 \\ 
			\midrule
			\multicolumn{13}{l}{\textit{\textbf{Span-level Hallucination Detection}}} 
			\\ \midrule
			\multicolumn{1}{c|}{Multi-View Attention 7B} & 41.5 & 49.6 & 35.7 & 50.6 & 38.5 & \textbf{73.7} & 55.2 & 53.5 & 57.1 & 50.7 & 45.1 & 57.9 \\

			\multicolumn{1}{c|}{\apples-CAPO-7B} & 50.9 & 64.4 & 42.3 & 56.4 & 57.1 & 56.5 & 60.4 & {67.1} & 54.9 & 56.9 & 62.4 & 52.3 \\
			\multicolumn{1}{c|}{\apples-CAPO-14B} & \textbf{57.6} & 64.2 & 52.3 & 54.8 & 52.5 & 57.3 & 62.6 & \underline{67.2} & 58.7 & 58.3 & 60.5 & 56.3 \\ 
			
			\multicolumn{1}{c|}{LettuceDetect-Base} & 36.7 & 53.0 & 28.1 & 61.5 & 62.7 & 60.4 & 57.4 & 58.2 & 56.6 & 55.4 & 59.4 & 52.0 \\ 
			\multicolumn{1}{c|}{LettuceDetect-Large} & 44.6 & 60.2 & 35.5 & \textbf{64.4} & \textbf{66.9} & 62.1 & 60.0 & 64.7 & 56.0 & 58.9 & {64.9} & 54.0 \\ 
			\multicolumn{1}{c|}{\textbf{Baseline-SFT-4B}} & 52.3 & 63.5 & 44.5 & 59.5 & 53.3 & 67.4 & 60.5 & 60.1 & 61.0 & 58.7 & 57.5 & 59.9 \\
			\multicolumn{1}{c|}{\textbf{Baseline-GRPO-4B}} & 53.2 & \textbf{70.9} & 42.6 & 59.3 & 62.8 & 56.1 & \underline{64.0} & \textbf{67.9} & 60.4 & 60.2 & \underline{66.4} & 55.1 \\
			\multicolumn{1}{c|}{\textbf{Baseline-CAPO-4B}} & 53.2 & 66.1 & 44.6 & 60.1 & 58.0 & 62.3 & {63.7} & 64.8 & 62.6 & \underline{60.4} & 62.2 & 58.7 \\

			\multicolumn{1}{c|}{\textbf{SOPO-4B}} & \underline{55.7} & \underline{70.1} & 46.2 & \underline{63.9} & \underline{66.4} & 61.6 & \textbf{66.0} & {67.1} & 65.0 & \textbf{63.3} & \textbf{67.3} & 59.8  \\
			\bottomrule
		\end{tabular}
		
	\end{adjustbox}
	\label{tab:main_table}
\end{table*}
\begin{table*}[ht!]
	\centering
	
	\caption{\textbf{Case-level hallucination detection results on RAGTruth.} 
		We compare our results with several approaches presented in Ragas \cite{es2025ragasautomatedevaluationretrieval}, Trulens, Luna \cite{belyi2024lunaevaluationfoundationmodel}, LettuceDetect \cite{kovács2025lettucedetecthallucinationdetectionframework}, Grounded in Context \cite{gerner2025oriongroundedcontextretrievalbased} and RAG-HAT \cite{song-etal-2024-rag}. Best scores are in bold, second best scores are underlined. 
	}
	\begin{adjustbox}{width=\linewidth}

		\begin{tabular}{c|ccc|ccc|ccc|ccc}
			\toprule
			\multicolumn{1}{c|}{\multirow{2}{*}{\textbf{Model}}} & \multicolumn{3}{c|}{\textbf{Summarization}} & \multicolumn{3}{c|}{\textbf{Question Answering}} & \multicolumn{3}{c|}{\textbf{Data-to-Text}} & \multicolumn{3}{c}{\textbf{Overall}} \\
			\multicolumn{1}{c|}{} & F1 & Precision & Recall & F1 & Precision & Recall & F1 & Precision & Recall & F1 & Precision & Recall \\ \midrule
			
			Ragas & 40.8 & 64.2 & 29.9 & 35.7 & 31.2 & 41.9 & 61.9 & 79.2 & 50.8 & 52.0 & 62.0 & 44.8 \\
			Trulens & 44.5 & 40.2 & 50.0 & 36.6 & 22.8 & \textbf{92.5} & 79.0 & 66.9 & \textbf{96.5} & 60.4 & 46.5 & 85.8 \\
			Luna & 52.5 & 40.0 & \underline{76.5} & 51.3 & 37.8 & 80.0 & 75.9 & 64.9 & 91.2 & 65.4 & 52.7 & \underline{86.1} \\
			LettuceDetect-Base & 50.5 & 53.9 & 47.6 & 65.5 & 60.6 & 71.3 & 87.9 & 89.3 & 86.5 & 76.1 & 76.6 & 75.5 \\
			LettuceDetect-Large  & 59.7 & 64.0 & 55.9 & 70.2 & 65.9 & 75.0 & 88.5 & 90.5 & 86.7 & 79.2 & 80.4 & 78.1 \\
			Grounded in Context  & \textbf{86.2} & \textbf{85.4} & \textbf{87.1} & \textbf{90.1} & \textbf{88.4} & \underline{91.9} & 72.2 & 64.4 & 82.2 & 83.1 & 79.4 & \textbf{87.1} \\
			RAG-HAT & 67.6 & 77.7 & 59.8 & \underline{74.8} & 76.5 & 73.1 & 91.6 & \textbf{92.9} & 90.3 & 83.9 & 87.3 & 80.8 \\
			
			\multicolumn{1}{c|}{\textbf{Baseline-SFT-4B}} & \underline{73.6} & 76.3 & 71.1 & 73.0 & 69.5 & 76.9 & 91.7 & 89.6 & \underline{93.8} & \underline{84.7} & 83.4 & 86.0 \\
			
			\multicolumn{1}{c|}{\textbf{Baseline-GRPO-4B}} & 69.5 & \underline{83.0} & 59.8 & 71.7 & 74.8 & 68.8 & 91.7 & \underline{92.5} & 91.0 & 84.0 & \underline{87.9} & 80.5 \\
			
			\multicolumn{1}{c|}{\textbf{Baseline-CAPO-4B}} & 70.5 & 76.1 & 65.7 & 71.8 & 66.0 & 78.8 & \underline{91.8} & 90.5 & 93.3 & 83.9 & 83.0 & 84.8 \\
			
			\multicolumn{1}{c|}{\textbf{SOPO-4B}} & 71.9 & 81.0 & 64.7 & 72.7 & \underline{78.8} & 67.5 & \textbf{92.0} & 92.2 & 91.9 & \textbf{84.8} & \textbf{88.0} & 81.9  \\
			\bottomrule
		\end{tabular}

	\end{adjustbox}
	\label{tab:main_table_case}
\end{table*}

\paragraph{Training Details.}Our RL framework is implemented using AREAL \cite{fu2025areallargescaleasynchronousreinforcement} along with dynamic sampling \cite{yu2025dapoopensourcellmreinforcement} to accelerate the training process. All experiments are conducted on 20 NVIDIA A800 GPUs. We employ a decoupled resource strategy, allocating 8 GPUs for rollouts, 8 for updates and 4 for feedback model inference. Full hyperparameters are listed in Table~\ref{tab:hyperparams}.

\paragraph{Inference.}
Consistent with the baselines, all methods (including SOPO) perform single-turn inference on the RAGTruth test set: given the retrieval-augmented context and the LLM-generated response, the model directly produces the reasoning chain and the predicted hallucination spans without any multi-turn interaction or teacher feedback. 

\section{Results}
\label{sec:results}
\subsection{Main Results}

\paragraph{Comprehensive Superiority on Span-level Detection.}

As presented in Table~\ref{tab:main_table}, SOPO (4B) establishes a new state-of-the-art in overall F1 score on the RAGTruth test set, significantly outperforming larger proprietary models (e.g., GPT-5, o3). Notably, it surpasses the previous best framework, LettuceDetect-Large, by \textbf{4.4 points} in overall F1 (63.3 vs. 58.9). SOPO also outperforms all baseline models trained by other methods (SFT, GRPO, and CAPO) across all three tasks (note that CAPO/RL4HS's original scores are calibrated for fair comparison, detailed in Appendix \ref{sec:calib_score}). Crucially, SOPO alleviates the aforementioned reward sparsity, maintaining robust performance even on highly constrained hard examples (Appendix~\ref{sec:hard_case}).

\paragraph{Generalization to Case-level Detection.}
Although optimized for span-level granularity, SOPO generalizes effectively to case-level detection. As shown in Table~\ref{tab:main_table_case}, it achieves a state-of-the-art Overall F1 of 84.8, surpassing dedicated methods (e.g., RAG-HAT) and all standard RL baselines. Notably, SOPO excels in Data-to-Text (92.0 F1) and yields the highest overall precision (88.0). 

\paragraph{Inference-time significance.}
We additionally evaluated the full SOPO checkpoint and the same-start CAPO and GRPO baselines over 10 full RAGTruth test-set inference runs per method. The mean overall span-F1 is 63.34 for SOPO, 59.75 for CAPO, and 59.49 for GRPO. The reported paired bootstrap differences are +3.59 points versus CAPO (nominal 95\% CI [1.52, 5.75], one-sided $p=0.0005$) and +3.85 versus GRPO ([1.33, 6.57], $p=0.0008$). For overall case-F1, the corresponding means are 84.65 versus 83.03 for both baselines, with nominal one-sided $p=0.0209$ and $p=0.0248$, respectively. The complete results and bootstrap protocol are reported in Appendix~\ref{sec:inference_significance}.

\paragraph{Breakthrough on Hard Examples.} 
Addressing the reward sparsity issue on difficult queries identified in Section \ref{intro}, SOPO demonstrates a significant breakthrough on hard examples. As detailed in Appendix~\ref{sec:hard_case}, SOPO consistently improves prediction scores across samples with extreme constraints, including overly short hallucinated spans, dense multi-span occurrences, and top-10\% longest contexts. By transforming sparse outcome rewards into dense Socratic guidance, SOPO successfully drives a robust positive shift in the score distribution compared to GRPO, directly contributing to its superior overall performance.

\paragraph{Out-of-Domain Evaluation.}
Across 10 inference runs on each of four out-of-domain (OOD) benchmarks, SOPO's mean primary score exceeds GRPO's: 73.90 vs. 72.51 span-F1 on PsiloQA, 89.02 vs. 85.68 span-F1 on Math-Shepherd-614, 75.93 vs. 72.55 mean balanced accuracy on LLM-AggreFact, and 69.77 vs. 67.52 accuracy on HaluEval. Detailed comparisons with CAPO and GRPO, including the paired-run uncertainty estimates for each benchmark, are provided in Appendix \ref{sec:additional_benchmarks}.

\paragraph{Accelerated Convergence and Autonomous Reasoning.}
\begin{figure*}[ht!]%
	\centering
	\begin{minipage}[t]{0.32\linewidth}
		\vspace{0pt}%
		\includegraphics[width=\linewidth]{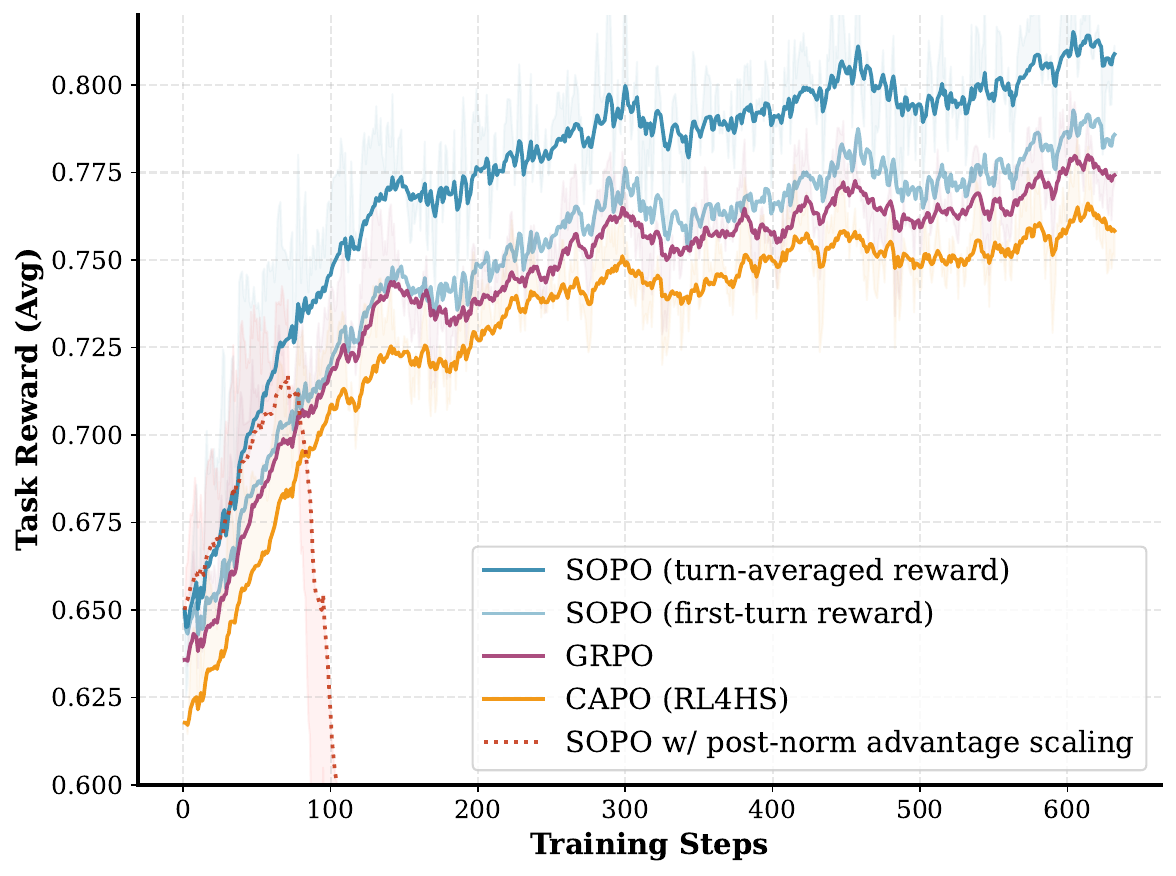}
		\caption{Mean task reward of RL over training steps.}
		\label{fig:reward}
	\end{minipage}
	\hfill
	\begin{minipage}[t]{0.32\linewidth}
		\vspace{0pt}%
		\includegraphics[width=\linewidth]{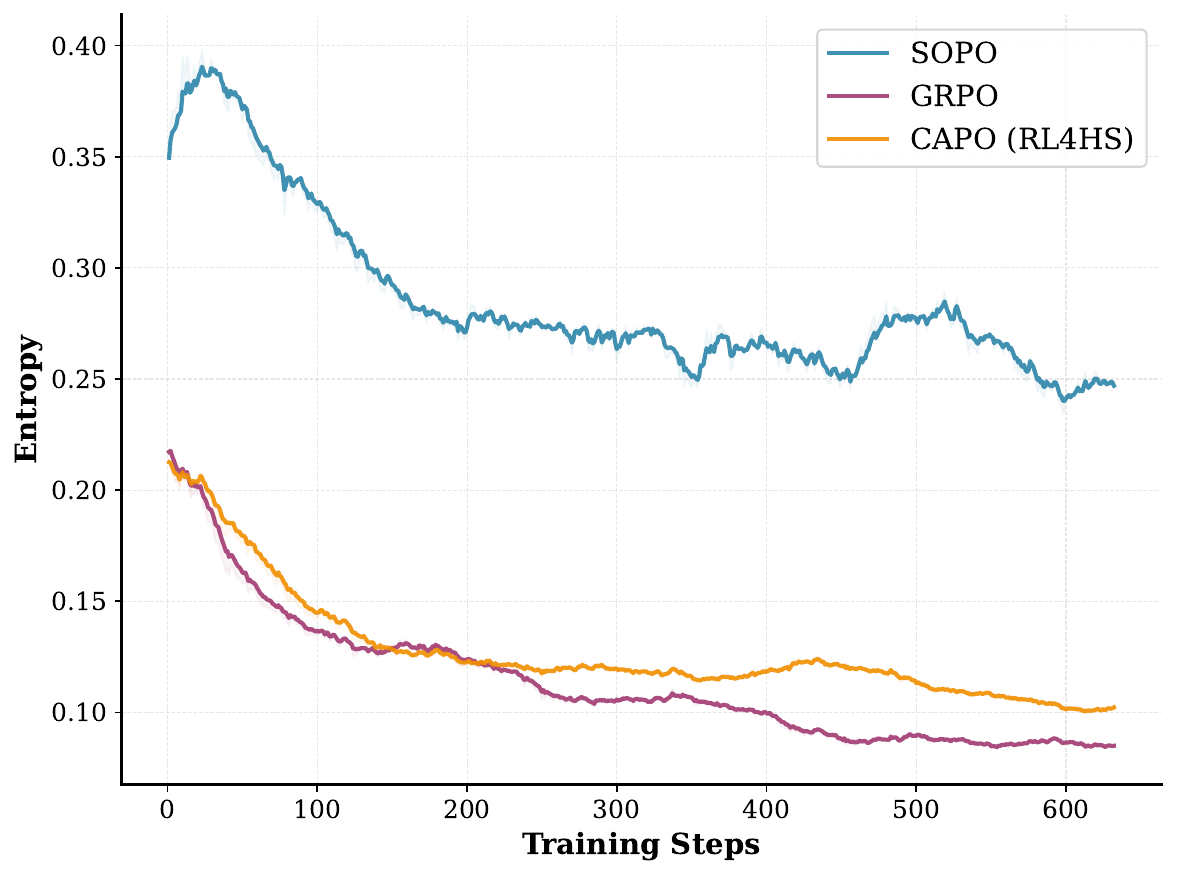}
		\caption{Entropy of RL over training steps.}
		\label{fig:entropy}
	\end{minipage}
	\hfill
	\begin{minipage}[t]{0.32\linewidth}
		\vspace{0pt}%
		\includegraphics[width=\linewidth]{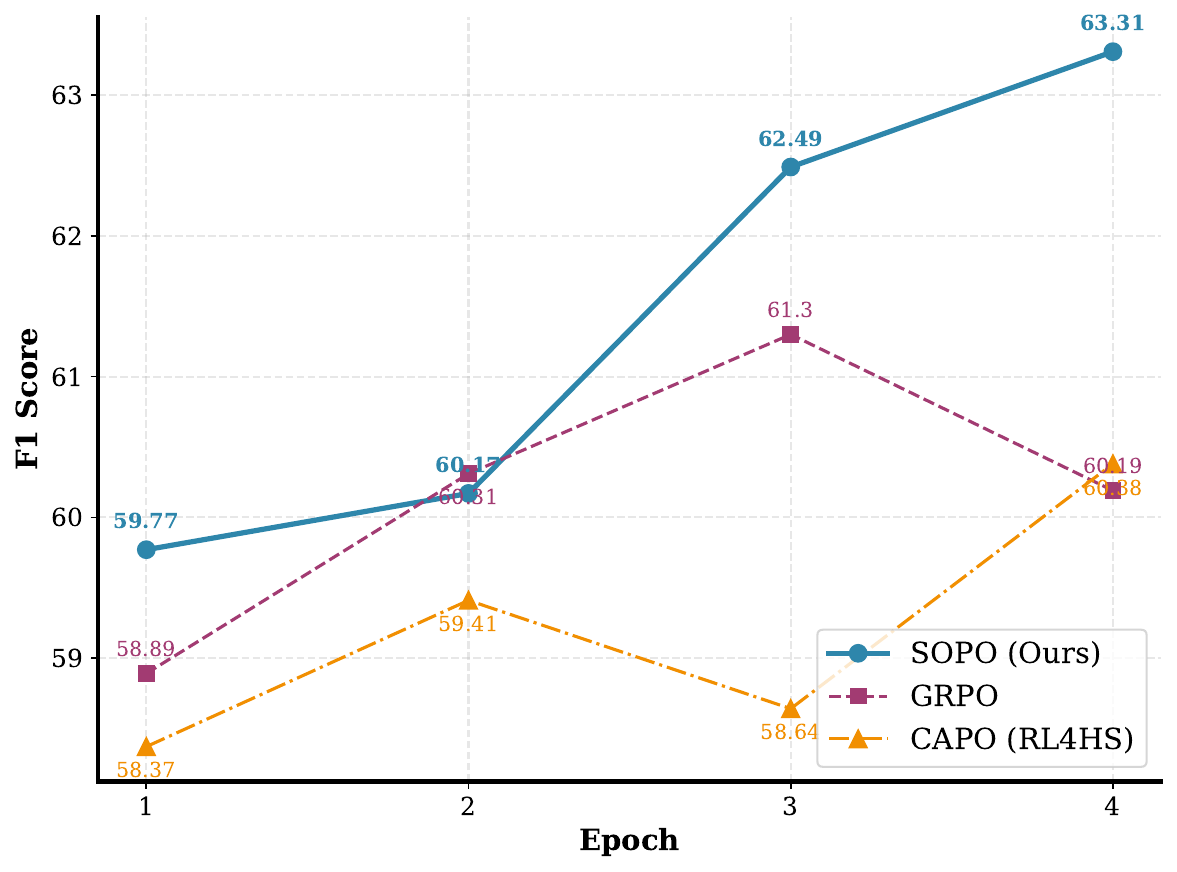}
		\caption{Overall F1 on RAGTruth test set over training epochs.}%
		\label{fig:convergence}%
	\end{minipage}
\end{figure*}

As illustrated in Figure~\ref{fig:reward}, SOPO demonstrates superior training efficiency, achieving a task reward at $\sim$200 steps that CAPO requires $\sim$600 steps to achieve. This rapid convergence is supported by evaluation results in Figure~\ref{fig:convergence}, where SOPO matches GRPO's peak performance (4 epochs) within just two epochs. Crucially, this efficiency stems from enhanced autonomous reasoning: as shown in Figure~\ref{fig:reward}, SOPO's performance in the first turn is consistently higher than other baselines even in the first epoch. This confirms that our turn-based advantage scaling mechanism successfully encourages the model to reduce reliance on teacher's feedback, driving it to perform accurate hallucination detection in the initial turn.

\paragraph{CoT Length Control}

Figure~\ref{fig:seqlen} highlights the effectiveness of our turn-based advantage scaling. While GRPO suffers from severe length inflation, SOPO maintains a stable length of $\sim$1,700 tokens per turn throughout training. Crucially, this stability coincides with monotonic performance gains, proving that SOPO improves reasoning quality by eliminating verbosity bias rather than simply generating redundant \myquote{filler} tokens \cite{shrivastava2025samplethinklessgroup}. As detailed in Appendix~\ref{sec:test_cost}, while SOPO introduces a modest one-time training FLOPs overhead (+18.9\%), the SOPO model reduces average token generation by 39.9\% and cuts inference latency by 47.0\%, yielding a highly practical framework for real-world applications.

\paragraph{Wider Policy Exploration.}

As illustrated in Figure~\ref{fig:entropy}, the policy entropy of SOPO remains consistently higher than that of both GRPO and CAPO throughout the training process. In terms of training dynamics, SOPO exhibits moderate entropy fluctuations even during the later stages (steps 450–600), indicating that the model retains the capacity for active exploration to further refine its performance. Conversely, GRPO and CAPO exhibit severe entropy collapse: their entropy values decay monotonically, approaching near-zero levels by step 600. This collapse indicates that the baseline policies have lost the ability to perform diverse sampling, resulting in performance plateauing where further training yields no additional gains.

\subsection{Feedback Scheme}
\label{sec:feedback_scheme}

\paragraph{Is Socratic guidance strictly necessary?} The complete Socratic feedback proposed in our method comprises two crucial components: step-wise analysis and guiding questions, which guide the student model to deduce the correct answer rather than directly revealing the ground truth. To verify the necessity of this specific design, we conducted experiments on different feedback schemes:
\begin{itemize}
\item \textbf{Analysis-only:} Providing step-wise analysis on the student's reasoning without the subsequent guiding question.
\item \textbf{Question-only:} Providing a guiding question directly without prior step-wise analysis.
\end{itemize}

\begin{figure*}[htbp]
	\centering
	
	\begin{minipage}[t]{0.31\linewidth}
		\vspace{0pt}
		\centering
		\includegraphics[width=\linewidth]{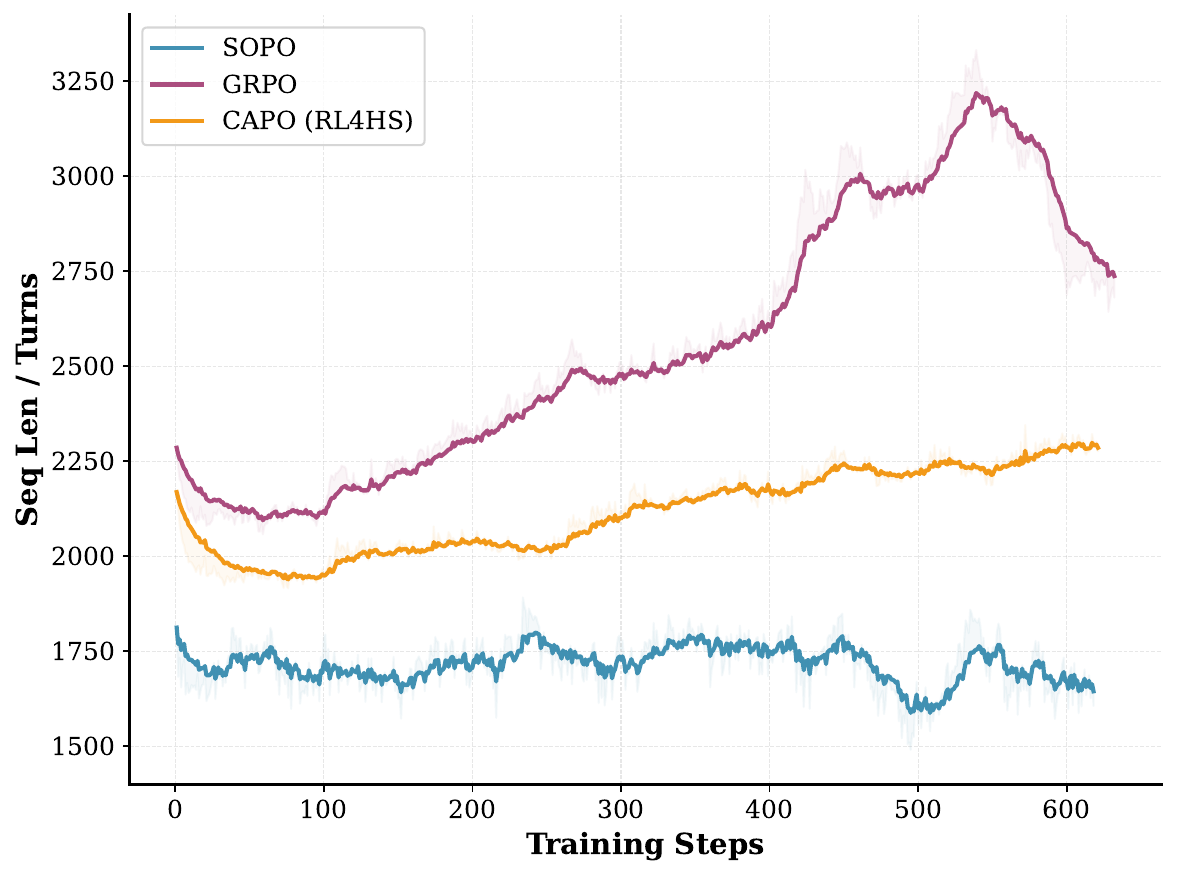}
		\caption{Sequence length per turn of RL.}
		\label{fig:seqlen}
	\end{minipage}%
	\hfill%
	\begin{minipage}[t]{0.31\linewidth}
		\vspace{0pt}%
		\centering
		\includegraphics[width=\linewidth]{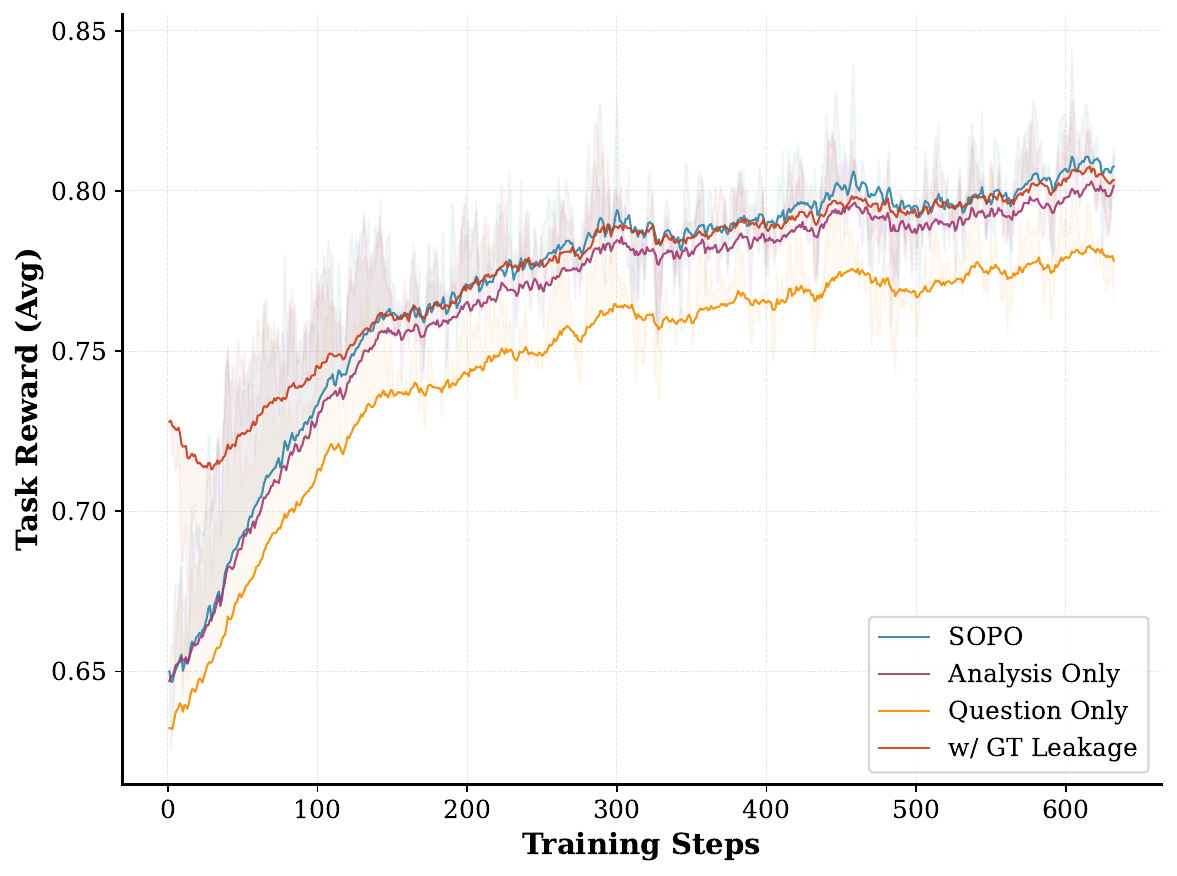}
		\caption{Mean reward of RL on different style feedback.}
		\label{fig:reward_socratic}
	\end{minipage}%
	\hfill%
	\begin{minipage}[t]{0.34\linewidth}
		\vspace{0pt}%
		\centering
		
		\makeatletter\def\@captype{table}\makeatother
		
		\caption{RAGTruth feedback-style comparison.}
		\label{tab:ablation_socratic}
		
		\small
		\setlength{\tabcolsep}{2pt}%
		\renewcommand{\arraystretch}{1.2}%
		
		\resizebox{\linewidth}{!}{
			\begin{tabular}{l c c c} 
				\toprule
				\textbf{Feedback Style} & \textbf{Span-F1} & \textbf{Prec.} & \textbf{Rec.} \\
				\midrule
				\textbf{SOPO} & \textbf{63.31} & 67.29 & \textbf{59.77} \\
				w/ GT Leakage & 61.65 & 64.43 & 59.09 \\
				Analysis-only & 59.49 & 64.57 & 55.16 \\
				Question-only & 62.25 & \textbf{68.25} & 57.22 \\
				Direct-Feedback & 60.81 & 64.41 & 57.59 \\
				\bottomrule
			\end{tabular}
		}
	\end{minipage}
\end{figure*}

As shown in Table~\ref{tab:ablation_socratic}, the model trained with the original SOPO achieves the best performance compared to other methods. Notably, the \myquote{Analysis-only} method yields the lowest score, whereas the \myquote{Question-only} method achieves results comparable to SOPO. This performance gap underscores the necessity of the guiding question component. It suggests that providing heuristic questions is more effective than providing passive critiques to improve the student model to precisely locate hallucination spans.

We also conduct experiments on \textbf{SOPO with ground truth leakage}, which provides both step-wise analysis and guiding question, but allowing explicit and implicit ground truth leakage. As illustrated in Figure~\ref{fig:reward_socratic}, Compared to the original SOPO, the \myquote{GT leakage} feedback results in inferior test performance, which exhibits higher rewards in the early training stage, but the original SOPO overtakes it after 120 steps. Crucially, while the final training rewards of both methods are nearly identical (gap $< 0.03$), Table~\ref{tab:ablation_socratic} reveals a significant performance drop (overall F1 gap $> 1.0$), indicating that the ground truth leakage encourages the model to overfit the RL training data, inflating the training reward scores while reducing the generalization ability. We also test \textbf{Direct-Feedback}, replacing Socratic guidance with directive, non-answer-revealing feedback (prompt in Appendix~\ref{sec:prompt}). It yields 60.81 span-F1 versus 63.31 for SOPO (Table~\ref{tab:ablation_socratic}).

\subsection{Model Scale and Heterogeneity}
\label{sec:feedback_scaling_teacher}
\paragraph{Teacher.}We evaluate teacher models across varying scales and architectures (Table~\ref{tab:feedback_scaling}). While scaling to the 80B teacher maximizes Overall F1 due to a precision improvement driven by highly accurate hallucination analysis, SOPO maintains strong performance even at smaller scales. Specifically, a 4B student guided by a 4B teacher still outperforms the GRPO baseline. Notably, the newer Qwen3-2507 4B model, known to surpass legacy 8B models on complex reasoning benchmarks (e.g., AIME25, LiveBench), also yields a higher F1 score than the legacy 8B teacher. This demonstrates that intrinsic reasoning capability is more critical than number of parameters. Furthermore, a heterogeneous student-teacher configuration using a different model family (GLM-4.7-Flash, 30BA3B) provides stronger guidance than a homogeneous setup with its Qwen3-30B-A3B counterpart.

\begin{table*}[t]%
	\centering
	
	\begin{minipage}[t]{0.48\linewidth}
		\centering
		\small
		\setlength{\tabcolsep}{3pt}%
		\caption{Performance of SOPO across diverse teacher model scales and architectures on the RAGTruth test set.}
		\label{tab:feedback_scaling}
		\begin{tabular}{lccc}
			\toprule
			\textbf{Teacher Model} & \textbf{F1} & \textbf{Prec.} & \textbf{Rec.} \\
			\midrule
			\shortstack[l]{Qwen3-Next-80B-A3B-Instruct} & \textbf{63.31} & \textbf{67.29} & 59.77 \\
			\shortstack[l]{GLM-4.7-Flash-30B-A3B}  & 62.48 & 65.23 & \textbf{59.96} \\
			\shortstack[l]{Qwen3-4B-Instruct-2507} & 62.16 & 66.49 & 58.35 \\
			\shortstack[l]{Qwen3-30B-A3B-Instruct-2507} & 61.67 & 63.78 & 59.69 \\
			Qwen3-8B & 60.47 & 65.51 & 56.14 \\
			\midrule
			\textit{\shortstack[l]{Baseline-GRPO-4B}} & {60.19} & {66.38} & {55.06} \\
			\bottomrule
		\end{tabular}
	\end{minipage}
	\hfill%
	\begin{minipage}[t]{0.48\linewidth}
		\centering
		\small
		\setlength{\tabcolsep}{3.5pt}%
		\caption{Performance of GRPO and SOPO across diverse student model scales on the RAGTruth test set.}
		\label{tab:student_scaling}
		\begin{tabular}{llccc}
			\toprule
			\textbf{Student Model} & \textbf{Method} & \textbf{F1} & \textbf{Prec.} & \textbf{Rec.} \\
			\midrule
			\multirow{2}{*}{Qwen3-1.7B} 
			& GRPO & 55.8 & \textbf{61.6} & 51.0 \\
			& SOPO & \textbf{57.7}& 61.2 & \textbf{54.5} \\
			\midrule
			\multirow{2}{*}{\shortstack[l]{Qwen3-4B-\\Thinking-2507}} 
			& GRPO & 60.2 & 66.4 & 55.1 \\
			& SOPO & \textbf{63.3} & \textbf{67.3} & \textbf{59.8} \\
			\midrule
			\multirow{2}{*}{Qwen3-8B} 
			& GRPO & 60.6 & 63.5 & \textbf{58.0} \\
			& SOPO & \textbf{61.0} & \textbf{66.2} & 56.5 \\
			\bottomrule
		\end{tabular}
	\end{minipage}
	
\end{table*}

\paragraph{Student.}To validate the robustness of student, we evaluate SOPO across diverse student capacities on RAGTruth (Table~\ref{tab:student_scaling}). SOPO consistently improves F1 across all scales, with the Qwen3-4B-2507 model yielding the largest gain (+3.1). The marginal improvements on the 1.7B and legacy 8B models stem from distinct bottlenecks: the 1.7B model lacks the long-context capacity required for long multi-turn trajectories, while the 8B model suffers from weaker intrinsic reasoning than the newer 2507 series, aligning with our teacher scaling results (Sec.~\ref{sec:feedback_scaling_teacher}).

\section{Comparison with Offline Chain-of-Thought Distillation}
\label{sec:distillation_main}
To distinguish online teacher feedback from offline distillation, we compare SOPO with 80B-CoT-Distill. SOPO queries a frozen Qwen3-Next-80B-A3B teacher for feedback conditioned on the student's evolving reasoning during on-policy RL rollouts and updates the student using trajectory rewards. By contrast, 80B-CoT-Distill uses the same teacher to generate chain-of-thought traces offline from RAGTruth training inputs and gold hallucination lists. These traces and gold lists form fixed supervised fine-tuning (SFT) targets for a student initialized from the same cold-start checkpoint; training uses 10,000 examples for approximately four epochs (457 steps). Thus, the baseline transfers teacher-generated reasoning through static targets rather than policy-conditioned, multi-turn feedback. Neither method uses the teacher or gold answers at inference.

\begin{table}[H]
    \centering
    \small
    \setlength{\tabcolsep}{6pt}
    \caption{Overall RAGTruth F1 (\%, mean $\pm$ standard deviation over 10 selected full-set, single-turn inference runs per checkpoint). All students share the cold-start initialization.}
    \label{tab:distill_main}
    \begin{tabular}{llcc}
        \toprule
        \textbf{Method} & \textbf{Training} & \textbf{Span-F1} & \textbf{Case-F1} \\
        \midrule
        SOPO & Interactive RL & $63.34\pm0.22$ & $84.65\pm0.21$ \\
        SOPO-FLOP & Interactive RL & $62.96\pm0.24$ & $83.72\pm0.32$ \\
        80B-CoT-Distill & Teacher-trace SFT & $50.09\pm0.77$ & $79.05\pm0.39$ \\
        \bottomrule
    \end{tabular}
\end{table}

As shown in Table~\ref{tab:distill_main}, SOPO exceeds 80B-CoT-Distill by 13.25 span-F1 points (nominal paired 95\% CI [10.26, 16.28]) and 5.60 case-F1 points ([3.73, 7.49]). The corresponding SOPO-FLOP differences are 12.87 ([9.77, 16.06]) and 4.67 ([2.69, 6.68]) points; all four one-sided $p$-values are $<10^{-4}$. SOPO-FLOP (531 steps) is approximately FLOP-matched to GRPO (Appendix~\ref{sec:inference_significance}), not to this SFT baseline.

These nominal bootstrap statistics follow Appendix~\ref{sec:inference_significance}.

\section{Conclusion}

We propose Socratic Guided Policy Optimization (SOPO) for robust span-level hallucination detection. By integrating Socratic guidance with Turn-based Advantage Scaling, SOPO resolves reward sparsity and ensures reasoning conciseness. Our 4B model achieves state-of-the-art performance on RAGTruth and exhibits strong generalization across diverse out-of-domain benchmarks. Crucially, this enhanced reasoning efficiency substantially reduces inference latency, presenting a highly potent and cost-effective framework for aligning small language models with complex verification tasks.

\section{Limitations}
\label{sec:limitations}

SOPO mitigates scalar reward instability via dense guidance, leaving advanced credit assignment techniques like integration with Process Reward Models (PRMs) for future work. Additionally, while our strict multi-turn verbosity penalty excels in verification tasks, it remains unclear whether this inadvertently suppresses the iterative trial-and-error exploration essential for open-ended generative domains like complex coding. Finally, repeated inference on fixed checkpoints provides bootstrap-based uncertainty estimates for the reported comparisons, but we have not assessed statistical significance across independent RL training seeds because full training runs are computationally costly. Nevertheless, SOPO's consistent margins across diverse benchmarks and model scales provide strong empirical confidence in its robustness.

\bibliographystyle{plain}
\bibliography{havor}

@misc{su2025learningreasonhallucinationspan,
      title={Learning to Reason for Hallucination Span Detection}, 
      author={Hsuan Su and Ting-Yao Hu and Hema Swetha Koppula and Kundan Krishna and Hadi Pouransari and Cheng-Yu Hsieh and Cem Koc and Joseph Yitan Cheng and Oncel Tuzel and Raviteja Vemulapalli},
      year={2025},
      eprint={2510.02173},
      archivePrefix={arXiv},
      primaryClass={cs.CL},
      url={https://arxiv.org/abs/2510.02173}, 
}

@misc{si2026faithlensdetectingexplainingfaithfulness,
      title={FaithLens: Detecting and Explaining Faithfulness Hallucination}, 
      author={Shuzheng Si and Qingyi Wang and Haozhe Zhao and Yuzhuo Bai and Guanqiao Chen and Kangyang Luo and Gang Chen and Fanchao Qi and Minjia Zhang and Baobao Chang and Maosong Sun},
      year={2026},
      eprint={2512.20182},
      archivePrefix={arXiv},
      primaryClass={cs.CL},
      url={https://arxiv.org/abs/2512.20182}, 
}

@misc{eliav2025clattercomprehensiveentailmentreasoning,
	title={CLATTER: Comprehensive Entailment Reasoning for Hallucination Detection}, 
	author={Ron Eliav and Arie Cattan and Eran Hirsch and Shahaf Bassan and Elias Stengel-Eskin and Mohit Bansal and Ido Dagan},
	year={2025},
	eprint={2506.05243},
	archivePrefix={arXiv},
	primaryClass={cs.CL},
	url={https://arxiv.org/abs/2506.05243}, 
}

@misc{niu2024ragtruthhallucinationcorpusdeveloping,
	title={RAGTruth: A Hallucination Corpus for Developing Trustworthy Retrieval-Augmented Language Models}, 
	author={Cheng Niu and Yuanhao Wu and Juno Zhu and Siliang Xu and Kashun Shum and Randy Zhong and Juntong Song and Tong Zhang},
	year={2024},
	eprint={2401.00396},
	archivePrefix={arXiv},
	primaryClass={cs.CL},
	url={https://arxiv.org/abs/2401.00396}, 
}

@misc{yang2025qwen3technicalreport,
	title={Qwen3 Technical Report}, 
	author={An Yang and Anfeng Li and Baosong Yang and Beichen Zhang and Binyuan Hui and Bo Zheng and Bowen Yu and Chang Gao and Chengen Huang and Chenxu Lv and Chujie Zheng and Dayiheng Liu and Fan Zhou and Fei Huang and Feng Hu and Hao Ge and Haoran Wei and Huan Lin and Jialong Tang and Jian Yang and Jianhong Tu and Jianwei Zhang and Jianxin Yang and Jiaxi Yang and Jing Zhou and Jingren Zhou and Junyang Lin and Kai Dang and Keqin Bao and Kexin Yang and Le Yu and Lianghao Deng and Mei Li and Mingfeng Xue and Mingze Li and Pei Zhang and Peng Wang and Qin Zhu and Rui Men and Ruize Gao and Shixuan Liu and Shuang Luo and Tianhao Li and Tianyi Tang and Wenbiao Yin and Xingzhang Ren and Xinyu Wang and Xinyu Zhang and Xuancheng Ren and Yang Fan and Yang Su and Yichang Zhang and Yinger Zhang and Yu Wan and Yuqiong Liu and Zekun Wang and Zeyu Cui and Zhenru Zhang and Zhipeng Zhou and Zihan Qiu},
	year={2025},
	eprint={2505.09388},
	archivePrefix={arXiv},
	primaryClass={cs.CL},
	url={https://arxiv.org/abs/2505.09388}, 
}

@misc{chaudhari2024rlhfdecipheredcriticalanalysis,
	title={RLHF Deciphered: A Critical Analysis of Reinforcement Learning from Human Feedback for LLMs}, 
	author={Shreyas Chaudhari and Pranjal Aggarwal and Vishvak Murahari and Tanmay Rajpurohit and Ashwin Kalyan and Karthik Narasimhan and Ameet Deshpande and Bruno Castro da Silva},
	year={2024},
	eprint={2404.08555},
	archivePrefix={arXiv},
	primaryClass={cs.LG},
	url={https://arxiv.org/abs/2404.08555}, 
}

@misc{feng2025dontwastemistakesleveraging,
	title={Don't Waste Mistakes: Leveraging Negative RL-Groups via Confidence Reweighting}, 
	author={Yunzhen Feng and Parag Jain and Anthony Hartshorn and Yaqi Duan and Julia Kempe},
	year={2025},
	eprint={2510.08696},
	archivePrefix={arXiv},
	primaryClass={cs.LG},
	url={https://arxiv.org/abs/2510.08696}, 
}

@misc{fatemi2025concisereasoningreinforcementlearning,
	title={Concise Reasoning via Reinforcement Learning}, 
	author={Mehdi Fatemi and Banafsheh Rafiee and Mingjie Tang and Kartik Talamadupula},
	year={2025},
	eprint={2504.05185},
	archivePrefix={arXiv},
	primaryClass={cs.CL},
	url={https://arxiv.org/abs/2504.05185}, 
}

@misc{shrivastava2025samplethinklessgroup,
	title={Sample More to Think Less: Group Filtered Policy Optimization for Concise Reasoning}, 
	author={Vaishnavi Shrivastava and Ahmed Awadallah and Vidhisha Balachandran and Shivam Garg and Harkirat Behl and Dimitris Papailiopoulos},
	year={2025},
	eprint={2508.09726},
	archivePrefix={arXiv},
	primaryClass={cs.CL},
	url={https://arxiv.org/abs/2508.09726}, 
}

@misc{shao2024deepseekmathpushinglimitsmathematical,
	title={DeepSeekMath: Pushing the Limits of Mathematical Reasoning in Open Language Models}, 
	author={Zhihong Shao and Peiyi Wang and Qihao Zhu and Runxin Xu and Junxiao Song and Xiao Bi and Haowei Zhang and Mingchuan Zhang and Y. K. Li and Y. Wu and Daya Guo},
	year={2024},
	eprint={2402.03300},
	archivePrefix={arXiv},
	primaryClass={cs.CL},
	url={https://arxiv.org/abs/2402.03300}, 
}

@misc{sun2025redeepdetectinghallucinationretrievalaugmented,
	title={ReDeEP: Detecting Hallucination in Retrieval-Augmented Generation via Mechanistic Interpretability}, 
	author={Zhongxiang Sun and Xiaoxue Zang and Kai Zheng and Yang Song and Jun Xu and Xiao Zhang and Weijie Yu and Yang Song and Han Li},
	year={2025},
	eprint={2410.11414},
	archivePrefix={arXiv},
	primaryClass={cs.CL},
	url={https://arxiv.org/abs/2410.11414}, 
}

@misc{zhou2024metacognitiveretrievalaugmentedlargelanguage,
	title={Metacognitive Retrieval-Augmented Large Language Models}, 
	author={Yujia Zhou and Zheng Liu and Jiajie Jin and Jian-Yun Nie and Zhicheng Dou},
	year={2024},
	eprint={2402.11626},
	archivePrefix={arXiv},
	primaryClass={cs.CL},
	url={https://arxiv.org/abs/2402.11626}, 
}

@misc{kovács2025lettucedetecthallucinationdetectionframework,
	title={LettuceDetect: A Hallucination Detection Framework for RAG Applications}, 
	author={Ádám Kovács and Gábor Recski},
	year={2025},
	eprint={2502.17125},
	archivePrefix={arXiv},
	primaryClass={cs.CL},
	url={https://arxiv.org/abs/2502.17125}, 
}

@misc{gerner2025oriongroundedcontextretrievalbased,
	title={ORION Grounded in Context: Retrieval-Based Method for Hallucination Detection}, 
	author={Assaf Gerner and Netta Madvil and Nadav Barak and Alex Zaikman and Jonatan Liberman and Liron Hamra and Rotem Brazilay and Shay Tsadok and Yaron Friedman and Neal Harow and Noam Bressler and Shir Chorev and Philip Tannor},
	year={2025},
	eprint={2504.15771},
	archivePrefix={arXiv},
	primaryClass={cs.LG},
	url={https://arxiv.org/abs/2504.15771}, 
}

@inproceedings{song-etal-2024-rag,
	title = "{RAG}-{HAT}: A Hallucination-Aware Tuning Pipeline for {LLM} in Retrieval-Augmented Generation",
	author = "Song, Juntong  and
	Wang, Xingguang  and
	Zhu, Juno  and
	Wu, Yuanhao  and
	Cheng, Xuxin  and
	Zhong, Randy  and
	Niu, Cheng",
	editor = "Dernoncourt, Franck  and
	Preo{\c{t}}iuc-Pietro, Daniel  and
	Shimorina, Anastasia",
	booktitle = "Proceedings of the 2024 Conference on Empirical Methods in Natural Language Processing: Industry Track",
	month = nov,
	year = "2024",
	address = "Miami, Florida, US",
	publisher = "Association for Computational Linguistics",
	url = "https://aclanthology.org/2024.emnlp-industry.113/",
	doi = "10.18653/v1/2024.emnlp-industry.113",
	pages = "1548--1558"
}

@misc{es2025ragasautomatedevaluationretrieval,
	title={Ragas: Automated Evaluation of Retrieval Augmented Generation}, 
	author={Shahul Es and Jithin James and Luis Espinosa-Anke and Steven Schockaert},
	year={2025},
	eprint={2309.15217},
	archivePrefix={arXiv},
	primaryClass={cs.CL},
	url={https://arxiv.org/abs/2309.15217}, 
}

@misc{belyi2024lunaevaluationfoundationmodel,
	title={Luna: An Evaluation Foundation Model to Catch Language Model Hallucinations with High Accuracy and Low Cost}, 
	author={Masha Belyi and Robert Friel and Shuai Shao and Atindriyo Sanyal},
	year={2024},
	eprint={2406.00975},
	archivePrefix={arXiv},
	primaryClass={cs.CL},
	url={https://arxiv.org/abs/2406.00975}, 
}

@misc{ogasa2025hallucinatedspandetectionmultiview,
	title={Hallucinated Span Detection with Multi-View Attention Features}, 
	author={Yuya Ogasa and Yuki Arase},
	year={2025},
	eprint={2504.04335},
	archivePrefix={arXiv},
	primaryClass={cs.CL},
	url={https://arxiv.org/abs/2504.04335}, 
}

@misc{fu2025areallargescaleasynchronousreinforcement,
	title={AReaL: A Large-Scale Asynchronous Reinforcement Learning System for Language Reasoning}, 
	author={Wei Fu and Jiaxuan Gao and Xujie Shen and Chen Zhu and Zhiyu Mei and Chuyi He and Shusheng Xu and Guo Wei and Jun Mei and Jiashu Wang and Tongkai Yang and Binhang Yuan and Yi Wu},
	year={2025},
	eprint={2505.24298},
	archivePrefix={arXiv},
	primaryClass={cs.LG},
	url={https://arxiv.org/abs/2505.24298}, 
}

@misc{yang2026grouprelativeadvantagebiased,
	title={Your Group-Relative Advantage Is Biased}, 
	author={Fengkai Yang and Zherui Chen and Xiaohan Wang and Xiaodong Lu and Jiajun Chai and Guojun Yin and Wei Lin and Shuai Ma and Fuzhen Zhuang and Deqing Wang and Yaodong Yang and Jianxin Li and Yikun Ban},
	year={2026},
	eprint={2601.08521},
	archivePrefix={arXiv},
	primaryClass={cs.LG},
	url={https://arxiv.org/abs/2601.08521}, 
}

@misc{dai2025stablereinforcementlearningefficient,
	title={Stable Reinforcement Learning for Efficient Reasoning}, 
	author={Muzhi Dai and Shixuan Liu and Qingyi Si},
	year={2025},
	eprint={2505.18086},
	archivePrefix={arXiv},
	primaryClass={cs.AI},
	url={https://arxiv.org/abs/2505.18086}, 
}

@misc{zhang2025edgegrpoentropydrivengrpoguided,
	title={EDGE-GRPO: Entropy-Driven GRPO with Guided Error Correction for Advantage Diversity}, 
	author={Xingjian Zhang and Siwei Wen and Wenjun Wu and Lei Huang},
	year={2025},
	eprint={2507.21848},
	archivePrefix={arXiv},
	primaryClass={cs.AI},
	url={https://arxiv.org/abs/2507.21848}, 
}

@misc{nan2025ngrponegativeenhancedgrouprelative,
	title={NGRPO: Negative-enhanced Group Relative Policy Optimization}, 
	author={Gongrui Nan and Siye Chen and Jing Huang and Mengyu Lu and Dexun Wang and Chunmei Xie and Weiqi Xiong and Xianzhou Zeng and Qixuan Zhou and Yadong Li and Xingzhong Xu},
	year={2025},
	eprint={2509.18851},
	archivePrefix={arXiv},
	primaryClass={cs.LG},
	url={https://arxiv.org/abs/2509.18851}, 
}

@misc{yu2025dapoopensourcellmreinforcement,
	title={DAPO: An Open-Source LLM Reinforcement Learning System at Scale}, 
	author={Qiying Yu and Zheng Zhang and Ruofei Zhu and Yufeng Yuan and Xiaochen Zuo and Yu Yue and Weinan Dai and Tiantian Fan and Gaohong Liu and Lingjun Liu and Xin Liu and Haibin Lin and Zhiqi Lin and Bole Ma and Guangming Sheng and Yuxuan Tong and Chi Zhang and Mofan Zhang and Wang Zhang and Hang Zhu and Jinhua Zhu and Jiaze Chen and Jiangjie Chen and Chengyi Wang and Hongli Yu and Yuxuan Song and Xiangpeng Wei and Hao Zhou and Jingjing Liu and Wei-Ying Ma and Ya-Qin Zhang and Lin Yan and Mu Qiao and Yonghui Wu and Mingxuan Wang},
	year={2025},
	eprint={2503.14476},
	archivePrefix={arXiv},
	primaryClass={cs.LG},
	url={https://arxiv.org/abs/2503.14476}, 
}

@misc{xu2026stepwisethinkcritiqueunifiedframework,
	title={Stepwise Think-Critique: A Unified Framework for Robust and Interpretable LLM Reasoning}, 
	author={Jiaqi Xu and Cuiling Lan and Xuejin Chen and Yan Lu},
	year={2026},
	eprint={2512.15662},
	archivePrefix={arXiv},
	primaryClass={cs.AI},
	url={https://arxiv.org/abs/2512.15662}, 
}

@misc{wu2025socraticrlnovelframework,
	title={Socratic RL: A Novel Framework for Efficient Knowledge Acquisition through Iterative Reflection and Viewpoint Distillation}, 
	author={Xiangfan Wu},
	year={2025},
	eprint={2506.13358},
	archivePrefix={arXiv},
	primaryClass={cs.AI},
	url={https://arxiv.org/abs/2506.13358}, 
}

@misc{rykov2025modelslielearnmultilingual,
	title={When Models Lie, We Learn: Multilingual Span-Level Hallucination Detection with PsiloQA}, 
	author={Elisei Rykov and Kseniia Petrushina and Maksim Savkin and Valerii Olisov and Artem Vazhentsev and Kseniia Titova and Alexander Panchenko and Vasily Konovalov and Julia Belikova},
	year={2025},
	eprint={2510.04849},
	archivePrefix={arXiv},
	primaryClass={cs.CL},
	url={https://arxiv.org/abs/2510.04849}, 
}

@misc{tang2024minicheckefficientfactcheckingllms,
	title={MiniCheck: Efficient Fact-Checking of LLMs on Grounding Documents}, 
	author={Liyan Tang and Philippe Laban and Greg Durrett},
	year={2024},
	eprint={2404.10774},
	archivePrefix={arXiv},
	primaryClass={cs.CL},
	url={https://arxiv.org/abs/2404.10774}, 
}

@misc{li2023haluevallargescalehallucinationevaluation,
	title={HaluEval: A Large-Scale Hallucination Evaluation Benchmark for Large Language Models}, 
	author={Junyi Li and Xiaoxue Cheng and Wayne Xin Zhao and Jian-Yun Nie and Ji-Rong Wen},
	year={2023},
	eprint={2305.11747},
	archivePrefix={arXiv},
	primaryClass={cs.CL},
	url={https://arxiv.org/abs/2305.11747}, 
}

@misc{wang2024mathshepherdverifyreinforcellms,
	title={Math-Shepherd: Verify and Reinforce LLMs Step-by-step without Human Annotations}, 
	author={Peiyi Wang and Lei Li and Zhihong Shao and R. X. Xu and Damai Dai and Yifei Li and Deli Chen and Y. Wu and Zhifang Sui},
	year={2024},
	eprint={2312.08935},
	archivePrefix={arXiv},
	primaryClass={cs.AI},
	url={https://arxiv.org/abs/2312.08935}, 
}

@misc{kaplan2020scalinglawsneurallanguage,
      title={Scaling Laws for Neural Language Models}, 
      author={Jared Kaplan and Sam McCandlish and Tom Henighan and Tom B. Brown and Benjamin Chess and Rewon Child and Scott Gray and Alec Radford and Jeffrey Wu and Dario Amodei},
      year={2020},
      eprint={2001.08361},
      archivePrefix={arXiv},
      primaryClass={cs.LG},
      url={https://arxiv.org/abs/2001.08361}, 
}

@misc{zhang2026critiquegrpoadvancingllmreasoning,
      title={Critique-GRPO: Advancing LLM Reasoning with Natural Language and Numerical Feedback},
      author={Xiaoying Zhang and Yipeng Zhang and Hao Sun and Kaituo Feng and Chaochao Lu and Chao Yang and Helen Meng},
      year={2026},
      eprint={2506.03106},
      archivePrefix={arXiv},
      primaryClass={cs.CL},
      url={https://arxiv.org/abs/2506.03106},
}

\newpage
\appendix
\appendix

\section{Broader Impacts}
\label{sec:impact_statement}

By enabling fine-grained hallucination localization, SOPO significantly enhances the safety and reliability of Retrieval-Augmented Generation (RAG) applications. This is particularly crucial in high-stakes domains, such as healthcare, legal analysis, and financial forecasting, where even minor factual errors can have severe consequences. To foster transparency and support further research in the community, we release our 1.7B and 4B models alongside the complete training codebase.

Despite these positive applications, the deployment of advanced hallucination detectors carries inherent risks. One primary concern is automation bias, where users may become overly reliant on the system and blindly trust its output, potentially missing subtle errors that the model fails to catch. Additionally, there is a risk of dual-use. Malicious actors could potentially use the precise diagnostic feedback provided by frameworks like SOPO to adversarially refine generated misinformation, making it harder for standard safety filters to detect.

To mitigate these risks, we strongly advocate that SOPO should be deployed strictly as an assistive diagnostic tool rather than an autonomous decision-maker. Maintaining a human-in-the-loop workflow is essential for final verification. Furthermore, language models inherently capture a static snapshot of information. As real-world knowledge naturally evolves and changes, the detector will require continuous updates and periodic retraining to maintain its reliability over time.

\section{Prompt}
\label{sec:prompt}
\begin{tcolorbox}[
	colback=white,%
	colframe=black,%
	coltitle=white,%
	title=\textbf{Prompt for RAGTruth Summarization}
	]
	{Below is the original document:}
	
	\{reference\}
	
	{Below is a summary of the document:}
	
	\{response\}
	
	{Your task is to determine whether the summary contains hallucinations.}
	{First, provide reasoning with the following format:}

	\#\# Step 1: $<$ your first reasoning step $>$

	\#\# Step 2: $<$ your next reasoning step $>$

	...(add as many steps as needed)
	Then, compile the labeled hallucinated spans into a JSON dict, with a key hallucination list and its value is a list of hallucinated spans. If there are potential hallucinations, the output should be in the following JSON format: {hallucination list: [hallucination span1, hallucination span2, ...]}. Otherwise, leave the value as an empty list as follows: {hallucination list: []}.
\end{tcolorbox}
\begin{tcolorbox}[
	colback=white,%
	colframe=black,%
	coltitle=white,%
	title=\textbf{Prompt for RAGTruth Question Answering}
	]
	Below is a question:
	
	\{question\}
	
	{Below are the related passages:}
	
	\{reference\}
	
	{Below is an answer:}
	
	\{response\}
	
	{Your task is to determine whether the answer contains hallucinations.}
	{First, provide reasoning with the following format:}

	\#\# Step 1: $<$ your first reasoning step $>$

	\#\# Step 2: $<$ your next reasoning step $>$

	...(add as many steps as needed)
	Then, compile the labeled hallucinated spans into a JSON dict, with a key hallucination list and its value is a list of hallucinated spans. If there are potential hallucinations, the output should be in the following JSON format: {hallucination list: [hallucination span1, hallucination span2, ...]}. Otherwise, leave the value as an empty list as follows: {hallucination list: []}.
\end{tcolorbox}
\begin{tcolorbox}[
	colback=white,%
	colframe=black,%
	coltitle=white,%
	title=\textbf{Prompt for RAGTruth Data-to-text}
	]
	Below is structured data in JSON format:
	
	\{reference\}
	
	Below is an overview article written in accordance with the structured data:
	
	\{response\}
	
	Your task is to determine whether the article contains hallucinations.
	First, provide reasoning with the following format:

	\#\# Step 1: $<$ your first reasoning step $>$

	\#\# Step 2: $<$ your next reasoning step $>$

	...(add as many steps as needed)
	Then, compile the labeled hallucinated spans into a JSON dict, with a key hallucination list and its value is a list of hallucinated spans. If there are potential hallucinations, the output should be in the following JSON format: {hallucination list: [hallucination span1, hallucination span2, ...]}. Otherwise, leave the value as an empty list as follows: {hallucination list: []}.
\end{tcolorbox}

\begin{tcolorbox}[
	colback=white,%
	colframe=black,%
	coltitle=white,%
	title=\textbf{Prompt for Socratic Guidance}
	]
	\# ROLE
	
	You are a Socratic teacher guiding a student to self-correct their logical reasoning. You must speak directly to the student using \myquote{you} and \myquote{your}.
	\newline
	
	\# INSTRUCTIONS
	
	Your task is to read the student's \myquote{CoT} and compare it against the \myquote{CONTEXT} and the \myquote{CORRECT\_ANSWER}. Find the exact step (e.g., \#\# Step 2) where the student makes their first logical error or hallucination.
	
	You must follow these strict rules:
	
	1. NEVER reveal the \myquote{CORRECT\_ANSWER} or tell the student the final True/False classification.
	
	2. Address the student directly (e.g., \myquote{You stated that...}).
	\newline
	
	\# HOW TO RESPOND
	
	Write your response in exactly two natural paragraphs:
	
	Paragraph 1 (Analysis): Directly tell the student which step contains the error. Briefly explain why their interpretation or logic in that specific step deviates from the provided CONTEXT. Do not use any special formatting or tags.
	
	Paragraph 2 (Guidance): Ask a leading Socratic question based on the CONTEXT to help the student realize their mistake. After your question, you MUST conclude your response with this exact sentence:
	
	\myquote{Please re-examine the logic in \#\# Step [n] and re-evaluate your conclusion from there.} (Replace [n] with the actual step number where the error occurred).
	\newline
	
	\# INPUT DATA
	
	\#\# CONTEXT
	
	\{context\_fmt\_str\}
	\newline
	
	\#\# STATEMENT
	
	\{statement\}
	\newline
	
	\#\# CORRECT\_ANSWER (HIDDEN TRUTH)
	
	\{correct\_answer\}
	\newline
	
	\#\# CoT (Student's Chain of Thought)
	
	\{result\}
\end{tcolorbox}

\begin{tcolorbox}[
	colback=white,%
	colframe=black,%
	coltitle=white,%
	title=\textbf{Prompt for Socratic Guidance w/ Ground Truth Leakage}
	]
	\# ROLE
	
	You are a Corrective Logic Guide helping a student fix their reasoning errors. You must speak directly to the student using \myquote{you} and \myquote{your}.
	\newline
	
	\# INSTRUCTIONS
	
	Your task is to read the student's \myquote{CoT} and compare it against the \myquote{CONTEXT} and the \myquote{CORRECT\_ANSWER}. Find the exact step (e.g., \#\# Step 2) where the student makes their first logical error or hallucination.
	
	You must follow these strict rules:
	
	1. LIMITED SPOILERS: You ARE ALLOWED to reveal specific facts from the CONTEXT or CORRECT\_ANSWER to correct a factual error in the student's current step.
	
	2. However, you MUST NOT simply blurt out the final classification (e.g., \myquote{The answer is False}) without explanation. Provide the evidence instead.
	
	3. Address the student directly (e.g., \myquote{You stated that...}).
	\newline
	
	\# HOW TO RESPOND
	
	Write your response in exactly two natural paragraphs:
	
	Paragraph 1 (Analysis): Directly tell the student which step contains the error. Explicitly contrast what they said versus what the CONTEXT actually states. It is okay to reveal the specific fact they got wrong. Do not use any special formatting or tags.
	
	Paragraph 2 (Guidance): Provide the specific evidence or correct fact to fix their mistake. After providing the fact, you MUST conclude your response with this exact sentence:
	
	\myquote{Please re-examine the logic in \#\# Step [n] using this correct information and re-evaluate your conclusion from there.} (Replace [n] with the actual step number where the error occurred).
	\newline
	
	\# INPUT DATA
	
	\#\# CONTEXT
	
	\{context\_fmt\_str\}
	\newline
	
	\#\# STATEMENT
	
	\{statement\}
	\newline
	
	\#\# CORRECT\_ANSWER
	
	\{correct\_answer\}
	\newline
	
	\#\# CoT (Student's Chain of Thought)
	
	\{result\}
\end{tcolorbox}

\begin{tcolorbox}[
	colback=white,
	colframe=black,
	coltitle=white,
	title=\textbf{Prompt for Direct-Feedback Ablation}
	]
	\small
	\# ROLE

	You give brief, non-Socratic, imperative feedback to a student who just failed a hallucination-detection attempt. This is an ablation baseline: feedback must be instructional, NOT analytical and NOT question-driven.
	\newline

	\# INPUT DATA

	\#\# CONTEXT

	\{context\_fmt\_str\}
	\newline

	\#\# STATEMENT

	\{statement\}
	\newline

	\#\# CORRECT\_ANSWER (HIDDEN TRUTH)

	\{correct\_answer\}
	\newline

	\#\# CoT (Student's Chain of Thought)

	\{result\}
	\newline

	\# CRITICAL RULES (NON-NEGOTIABLE)

	1. Output ONLY 1-2 short imperative sentences of direct instruction feedback.

	2. Do NOT use Socratic questions (no \myquote{why...}, \myquote{what if...}, \myquote{can you re-check...?}).

	3. Do NOT write step-wise analysis, critique essays, or \verb|<analysis>/<guidance>| tags.

	4. Do NOT reveal the CORRECT\_ANSWER contents, gold spans, or the final True/False label.

	5. Do NOT quote long evidence from CONTEXT. Keep the hint coarse and actionable.

	6. Prefer terse instructional phrasing similar to:

	- \myquote{Some information is missing. Please revise your hallucination list.}

	- \myquote{Please check the numbers again.}

	- \myquote{There is a factual conflict. Re-check your conclusion.}

	- \myquote{A key detail is unsupported. Please inspect the statement again.}

	7. You may lightly adapt the instruction to the error type (missing info / numeric mismatch / conflict / unsupported claim), but stay brief and imperative.
	\newline

	\# OUTPUT

	Return ONLY the short instruction text. No preamble, no markdown headings, no lists.
\end{tcolorbox}

\section{Training Details}
\label{sec:training_details}

The code for our experiments is available at \url{https://github.com/patsnap/SOPO} and the hyperparameters are listed in Table \ref{tab:hyperparams}.

\begin{table}[h]
	\centering
	\small%
	\renewcommand{\arraystretch}{1.2}%
	\caption{Hyperparameter of SFT and RL training.}
	\label{tab:hyperparams}
	\vspace{2mm}
	
	\begin{tabular}{lcc}
		\toprule
		\textbf{Hyperparameter} & \textbf{SFT (Cold-start)} & \textbf{RL} \\
		\midrule
		Learning Rate & $7\times 10^{-6}$ & $1\times 10^{-6}$ \\
		Batch Size & 16 & 64 \\
		Epochs & 3 & 4 \\
		Warmup Epoch & 1 & 1 \\
		LR Scheduler & Cosine & Constant \\
		Optimizer & Adam & Adam \\
		Adam $\beta_1$ & 0.9 & 0.9 \\
		Adam $\beta_2$ & 0.95 & 0.999 \\
		Weight Decay & 0.01 & 0.05 \\
		Group Size ($G$) & - & 8 \\
		Rollout Temperature & - & 1.0 \\
		Rollout Top-p & - & 0.99 \\
		Max Tokens Per Traj & - & 32768 \\
		Max Tokens Per Turn & - & 8192 \\
		Max Turns ($T$) & - & 6 \\
		Advantage Penalty Per Turn ($p_{\text{penalty}}$) & - & 0.2 \\
		Advantage Norm Level & - & Group \\
		Feedback Reward Threshold ($\delta$) & - & 0.6 \\
		Prob Feedback ($p_{\text{feedback}}$) & - & 0.6 \\
		\bottomrule
	\end{tabular}
\end{table}

\section{Computational Cost}
\label{sec:cost_analysis}

In this section, we provide a comprehensive breakdown of the computational costs associated with SOPO, addressing both training-time FLOPs and test-time inference latency. To ensure a strictly matched, unbiased evaluation, we compare both methods under the Dynamic Sampling (DS) configuration.

\subsection{FLOPs Calculation}
We estimate the total training floating-point operations (FLOPs) using standard scaling laws \cite{kaplan2020scalinglawsneurallanguage}, approximating the cost as $2ND$ for forward generation and $6ND$ for backward updates, where $N$ is the number of active parameters and $D$ is the number of processed tokens. 

Let $T$ denote the total training trajectories and $\rho$ denote the dynamic sampling rate. For the GRPO baseline under the DS configuration, the student model must first generate all trajectories before filtering for updates. Thus, the total FLOPs are split into full forward generation and sampled backward updates:
\begin{equation}
	\label{eq:grpo_flops}
	\mathcal{F}_{\text{GRPO}} = \underbrace{2N_{\text{stu}}D_{\text{grpo}}T}_{\text{Forward: Full } T} + \underbrace{6N_{\text{stu}}D_{\text{grpo}}(\rho \cdot T)}_{\text{Backward: Only } \rho \text{ of } T}
\end{equation}

For SOPO, forward generation is executed for all trajectories ($100\%$ of $T$) using both the student ($N_{\text{stu}}$) and the frozen MoE teacher ($N_{\text{tea}}$, 3B active parameters). However, the backward update is only performed on a filtered fraction of trajectories ($\rho$ of $T$) using exclusively the student model. The SOPO FLOPs formulation is:

\begin{equation}
	\label{eq:sopo_flops}
	\mathcal{F}_{\text{SOPO}} = \underbrace{2N_{\text{stu}}D_{\text{stu}}T + 2N_{\text{tea}}D_{\text{tea}}T}_{\text{Forward: Full } T} + \underbrace{6N_{\text{stu}}D_{\text{sopo}}(\rho \cdot T)}_{\text{Backward: Only } \rho \text{ of } T}
\end{equation}
\subsection{Empirical Cost Analysis}
To provide a strictly matched computational baseline, we evaluate both models using Dynamic Sampling with $T=320,000$.

\begin{table}[H]
	\centering
	\small
	\caption{Trajectory statistics.}
	\label{tab:stat_with_ds}
	\begin{tabular}{lccc}
		\toprule
		\textbf{Statistic} & \textbf{GRPO} & \textbf{SOPO} & \textbf{Difference} \\
		\midrule
		Generation Length (multi-turn sum) & 2700.1 & 2529.4 & -6.3\% \\
		Dynamic Sampling Rate ($\rho$) & 41.30\% & 48.02\% &  +6.7\% \\
		Turns Per Trajectory & 1 & 1.57 & - \\
		Teacher Gen Length Per Feedback & - & 172.8 & - \\
		Student Gen Length Per Turn & 2700.1 & 1534.4 & -43.2\% \\
		\bottomrule
	\end{tabular}
\end{table}

\begin{table}[H]
	\centering
	\small
	\caption{FLOPs calculation (ExaFLOPs).}
	\label{tab:flops_with_ds}
	\begin{tabular}{llccc}
		\toprule
		\textbf{Training Phase} & \textbf{Component} & \textbf{GRPO} & \textbf{SOPO} & \textbf{Difference} \\
		\midrule
		\multirow{2}{*}{Forward ($2ND$)} & Student & 9.37 & 8.62 & -8.0\% \\
		& Teacher & 0.00 & 3.46 &  - \\ 
		\midrule
		Backward ($6ND$) & Student & 11.61 & 12.87 & +10.6\% \\
		\midrule
		\textbf{Total} & & 20.98 & 24.95 & +18.9\% \\
		\bottomrule
	\end{tabular}
\end{table}

\begin{table}[H]
	\centering
	\small
	\caption{Wall-clock time and GPU hours.}
	\label{tab:time_with_ds}
	\begin{tabular}{lccc}
		\toprule
		\textbf{Metric} & \textbf{GRPO} & \textbf{SOPO} & \textbf{Difference} \\
		\midrule
		Training Time (Hours) & 17.88 & 15.82 & -11.5\% \\
		GPU Hours (NVIDIA-A800) & 286.08 & 316.40 & +10.6\% \\
		\bottomrule
	\end{tabular}
\end{table}

As detailed in Table~\ref{tab:flops_with_ds}, SOPO introduces an 18.9\% FLOPs overhead due to the decoupled teacher forward pass. Consequently, SOPO requires 10.6\% more total GPU hours (Table~\ref{tab:time_with_ds}), utilizing 20 GPUs: 16 for RL training and 4 for teacher deployment. Notably, SOPO yields a student model with a 43.2\% reduction in single-turn response length. This enhanced reasoning efficiency during training rollouts effectively offsets the teacher's invocation overhead, ultimately reducing the absolute wall-clock training time by 11.5\%.

Furthermore, the actual inference overhead of the teacher is lower due to Automatic Prefix Caching in modern LLM inference frameworks (e.g., SGLang, vLLM). Teacher prompts contain a relatively long shared prefix, approximately 40\% of the teacher's input, KV-cache is computed only once and reuses it across all rollouts.

\subsection{Training Efficiency and Convergence}

Beyond the reduction in generation length, the practical efficiency of SOPO is fundamentally rooted in its superior convergence rate. The dense Socratic guidance significantly reduces exploration variance, allowing SOPO to achieve the peak performance of GRPO in just 50\% of the total training steps.

\subsection{Test-Time Cost and Deployment Latency}
\label{sec:test_cost}

\begin{figure}[H]
	\centering
	\begin{minipage}[t]{0.55\linewidth}
		\vspace{0pt}%
		\makeatletter\def\@captype{table}\makeatother%
		\centering
		\small
		\setlength{\tabcolsep}{3pt}%
		\caption{Average inference latency and token counts during testing on RAGTruth (Concurrency = 32).}
		\label{tab:test_latency}
		\begin{tabular}{llccc}
			\toprule
			\textbf{Method} &  \textbf{Latency Avg (ms)} & \textbf{Latency Std} & \textbf{\# Token Avg} \\
			\midrule
			GRPO &  478.2 & 225.1 & 1250.2 \\
			SOPO &  \textbf{253.6 (-47.0\%)} & 81.8 & \textbf{750.9 (-39.9\%)} \\
			\bottomrule
		\end{tabular}
	\end{minipage}
	\hfill
	\begin{minipage}[t]{0.4\linewidth}
		\vspace{0pt}%
		\centering
		\includegraphics[width=\linewidth]{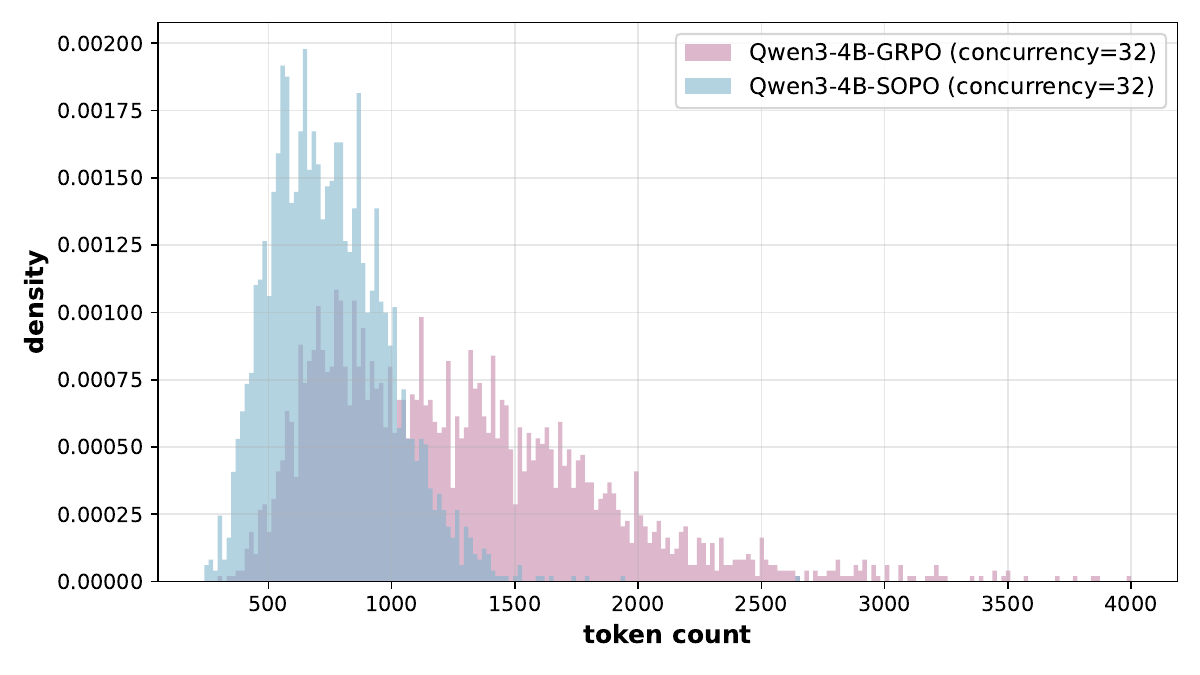}
		\caption{Output Length Distributions of SOPO and GRPO on RAGTruth.}
		\label{fig:output_dist}
	\end{minipage}
\end{figure}

The modest training overhead translates into massive efficiency gains during deployment. As shown in Table~\ref{tab:test_latency} and Figure \ref{fig:output_dist}, the SOPO student model naturally suppresses verbosity, reducing average token counts by 39.9\% and cutting inference latency by {47.0\%}. Trading a one-time 18.9\% training overhead for a 47\% reduction in real-world deployment latency represents a highly practical and significant contribution to the community.

\section{Inference-Time Significance on RAGTruth}
\label{sec:inference_significance}

\paragraph{Compute-matched checkpoints.}
Let $S=632$ be the full-training step count for both GRPO and SOPO. Under the dynamic-sampling configuration in Appendix~\ref{sec:cost_analysis}, let $F_m$ and $H_m$ denote the full-budget FLOPs and GPU-hours for method $m$, respectively. Table~\ref{tab:flops_with_ds} gives $(F_{\mathrm{GRPO}},F_{\mathrm{SOPO}})=(20.98,24.95)$ ExaFLOPs for $T=320{,}000$ trajectories, and Table~\ref{tab:time_with_ds} gives $(H_{\mathrm{GRPO}},H_{\mathrm{SOPO}})=(286.08,316.40)$ GPU-hours. Approximating the SOPO cost per step by $F_{\mathrm{SOPO}}/S$ or $H_{\mathrm{SOPO}}/S$, we choose the nearest integer step counts that match GRPO's full-budget costs:
\begin{align}
    s_{\mathrm{FLOP}}
    &= \operatorname{round}\!\left(\frac{F_{\mathrm{GRPO}}}{F_{\mathrm{SOPO}}/S}\right) \notag\\
    &= \operatorname{round}\!\left(\frac{632\times20.98}{24.95}\right)
     = 531, \label{eq:sopo_flop_checkpoint}\\
    s_{\mathrm{GPUh}}
    &= \operatorname{round}\!\left(\frac{H_{\mathrm{GRPO}}}{H_{\mathrm{SOPO}}/S}\right) \notag\\
    &= \operatorname{round}\!\left(\frac{632\times286.08}{316.40}\right)
     = 571. \label{eq:sopo_gpuh_checkpoint}
\end{align}
Equation~\eqref{eq:sopo_flop_checkpoint} defines SOPO-FLOP, and Eq.~\eqref{eq:sopo_gpuh_checkpoint} defines SOPO-GPUh. Using average per-step costs, these checkpoints approximately match GRPO's 632-step FLOP and GPU-hour budgets, respectively.

For each checkpoint and baseline, we evaluate 10 full-set inference runs on the fixed 2,700-example RAGTruth test set. Tables~\ref{tab:main_table} and~\ref{tab:main_table_case} report single-run scores; Table~\ref{tab:ragtruth_significance} summarizes the repeated-inference results. Differences are computed before rounding the displayed means.

\begin{table*}[t]
    \centering
    \small
    \setlength{\tabcolsep}{4pt}
    \caption{Repeated-inference RAGTruth results (\%, 10 runs per method). Nominal paired-bootstrap 95\% CIs and one-sided $p$-values use 10,000 replicates (seed 42), drawing one run per pool and resampling matched test examples. Pools were selected by span-F1 (including baseline runs from multiple evaluation campaigns); case-F1 uses the same runs. Estimates are conditional on these pools and do not measure training-seed variability.}
    \label{tab:ragtruth_significance}
    \begin{tabular}{clrrrcr}
        \toprule
        \textbf{SOPO checkpoint} & \textbf{Baseline} & \textbf{SOPO mean} & \textbf{Base mean} & $\boldsymbol{\Delta}$ \textbf{F1} & \textbf{95\% CI for $\Delta$} & \textbf{One-sided $p$} \\
        \midrule
        \multicolumn{7}{l}{\emph{Overall span-F1}} \\
        \multirow{2}{*}{SOPO} & CAPO & \multirow{2}{*}{63.34} & 59.75 & +3.59 & [1.52, 5.75] & 0.0005 \\
        & GRPO & & 59.49 & +3.85 & [1.33, 6.57] & 0.0008 \\
        \midrule
        \multirow{2}{*}{SOPO-FLOP} & CAPO & \multirow{2}{*}{62.96} & 59.75 & +3.21 & [0.98, 5.38] & 0.0031 \\
        & GRPO & & 59.49 & +3.47 & [0.93, 6.00] & 0.0039 \\
        \midrule
        \multirow{2}{*}{SOPO-GPUh} & CAPO & \multirow{2}{*}{62.74} & 59.75 & +2.99 & [0.95, 5.07] & 0.0016 \\
        & GRPO & & 59.49 & +3.25 & [0.80, 5.60] & 0.0057 \\
        \midrule
        \multicolumn{7}{l}{\emph{Overall case-F1}} \\
        \multirow{2}{*}{SOPO} & CAPO & \multirow{2}{*}{84.65} & 83.03 & +1.61 & [0.07, 3.07] & 0.0209 \\
        & GRPO & & 83.03 & +1.61 & [0.00, 3.25] & 0.0248 \\
        \midrule
        \multirow{2}{*}{SOPO-FLOP} & CAPO & \multirow{2}{*}{83.72} & 83.03 & +0.69 & [-1.00, 2.37] & 0.2189 \\
        & GRPO & & 83.03 & +0.69 & [-1.07, 2.44] & 0.2293 \\
        \midrule
        \multirow{2}{*}{SOPO-GPUh} & CAPO & \multirow{2}{*}{84.07} & 83.03 & +1.04 & [-0.56, 2.64] & 0.0990 \\
        & GRPO & & 83.03 & +1.04 & [-0.60, 2.75] & 0.1138 \\
        \bottomrule
    \end{tabular}
\end{table*}

At the checkpoints approximately matched to GRPO's FLOP and GPU-hour budgets, SOPO improves overall span-F1 over GRPO by 3.47 and 3.25 points, respectively, and over CAPO by 3.21 and 2.99 points. All four nominal paired 95\% bootstrap intervals exclude zero (one-sided $p\leq 0.0057$; Table~\ref{tab:ragtruth_significance}).

\section{Evaluation on Additional Benchmarks}
\label{sec:additional_benchmarks}

We evaluate the full-training SOPO checkpoint and its FLOP-matched checkpoint (SOPO-FLOP, 531 steps) on four out-of-domain (OOD) benchmarks using their primary metrics: span-F1 for PsiloQA and Math-Shepherd-614, the unweighted mean balanced accuracy (BAcc) for LLM-AggreFact, and overall accuracy for HaluEval. Table~\ref{tab:ood_means} presents means and standard deviations across 10 inference runs per method; Table~\ref{tab:ood_bootstrap} presents nominal sample-paired bootstrap comparisons. The GPU-hour-matched checkpoint is not included in the OOD evaluation.

\begin{table*}[t]
    \centering
    \footnotesize
    \setlength{\tabcolsep}{3pt}
    \caption{OOD primary-metric means $\pm$ standard deviations (\%) over 10 inference runs per checkpoint or baseline group.}
    \label{tab:ood_means}
    \begin{tabular}{lcccc}
        \toprule
        \textbf{Checkpoint} & \textbf{PsiloQA span-F1} & \textbf{Math-Shepherd span-F1} & \textbf{AggreFact mean BAcc} & \textbf{HaluEval accuracy} \\
        \midrule
        SOPO & $73.90\pm0.36$ & $89.02\pm0.49$ & $75.93\pm0.15$ & $69.77\pm0.03$ \\
        SOPO-FLOP & $74.17\pm0.20$ & $87.59\pm0.62$ & $75.23\pm0.10$ & $69.23\pm0.06$ \\
        CAPO & $72.67\pm0.18$ & $85.89\pm0.60$ & $75.29\pm0.11$ & $69.19\pm0.19$ \\
        GRPO & $72.51\pm0.17$ & $85.68\pm0.61$ & $72.55\pm0.90$ & $67.52\pm0.35$ \\
        \bottomrule
    \end{tabular}
\end{table*}

\begin{table*}[t]
    \centering
    \footnotesize
    \setlength{\tabcolsep}{4pt}
    \caption{OOD primary-metric sample-paired bootstrap comparisons (percentage points). Reported $\Delta$, 95\% CIs, and one-sided $p$-values pair the median-scoring run from each method's 10 inference runs and use $B=10{,}000$ paired example-resampling replicates (seed 42).}
    \label{tab:ood_bootstrap}
    \begin{tabular}{cclrcr}
        \toprule
        \textbf{Benchmark} & \textbf{SOPO checkpoint} & \textbf{Baseline} & $\boldsymbol{\Delta}$ & \textbf{95\% CI for $\Delta$} & \textbf{One-sided $p$} \\
        \midrule
        \multirow{4}{*}{PsiloQA} & \multirow{2}{*}{SOPO} & CAPO & +1.28 & $[-0.26, 2.92]$ & 0.0524 \\
        & & GRPO & +1.47 & $[-0.05, 3.12]$ & 0.0282 \\
        \cmidrule{2-6}
        & \multirow{2}{*}{SOPO-FLOP} & CAPO & +1.48 & $[-0.15, 3.19]$ & 0.0353 \\
        & & GRPO & +1.66 & $[0.08, 3.43]$ & 0.0212 \\
        \midrule
        \multirow{4}{*}{Math-Shepherd} & \multirow{2}{*}{SOPO} & CAPO & +3.04 & $[0.35, 5.99]$ & 0.0131 \\
        & & GRPO & +3.19 & $[-0.03, 6.73]$ & 0.0267 \\
        \cmidrule{2-6}
        & \multirow{2}{*}{SOPO-FLOP} & CAPO & +1.39 & $[-1.79, 4.64]$ & 0.1950 \\
        & & GRPO & +1.53 & $[-2.48, 5.60]$ & 0.2220 \\
        \midrule
        \multirow{4}{*}{LLM-AggreFact} & \multirow{2}{*}{SOPO} & CAPO & +0.58 & $[-0.32, 1.46]$ & 0.0994 \\
        & & GRPO & +2.73 & $[1.84, 3.64]$ & $<10^{-4}$ \\
        \cmidrule{2-6}
        & \multirow{2}{*}{SOPO-FLOP} & CAPO & -0.12 & $[-1.06, 0.81]$ & 0.5950 \\
        & & GRPO & +2.03 & $[1.13, 2.97]$ & $<10^{-4}$ \\
        \midrule
        \multirow{4}{*}{HaluEval} & \multirow{2}{*}{SOPO} & CAPO & +0.46 & $[0.18, 0.75]$ & 0.0005 \\
        & & GRPO & +2.07 & $[1.76, 2.37]$ & $<10^{-4}$ \\
        \cmidrule{2-6}
        & \multirow{2}{*}{SOPO-FLOP} & CAPO & -0.12 & $[-0.41, 0.18]$ & 0.7740 \\
        & & GRPO & +1.49 & $[1.19, 1.79]$ & $<10^{-4}$ \\
        \bottomrule
    \end{tabular}
\end{table*}

\subsection{Span-Level: PsiloQA} 
PsiloQA \cite{rykov2025modelslielearnmultilingual} evaluates fine-grained QA on its English subset using span-F1. SOPO averages 73.90 over 10 inference runs, compared with 72.67 for CAPO and 72.51 for GRPO; SOPO-FLOP averages 74.17 (Table~\ref{tab:ood_means}).

\subsection{Case-Level: LLM-AggreFact} 
LLM-AggreFact \cite{tang2024minicheckefficientfactcheckingllms} aggregates grounded factuality datasets; we exclude its RAGTruth subset and report the unweighted mean balanced accuracy across the remaining datasets. SOPO averages 75.93 versus 75.29 for CAPO and 72.55 for GRPO. Its nominal paired 95\% interval versus GRPO is positive (Table~\ref{tab:ood_bootstrap}).

\subsection{Case-Level: HaluEval} 
On HaluEval \cite{li2023haluevallargescalehallucinationevaluation}, we compare overall binary accuracy. SOPO averages 69.77 over 10 inference runs, versus 69.19 for CAPO and 67.52 for GRPO; its nominal paired 95\% intervals are above zero against both baselines (Table~\ref{tab:ood_bootstrap}).

\subsection{Out-of-Domain: Math-Shepherd} 
To construct Math-Shepherd-614, we randomly sample 1,000 instances from the public Math-Shepherd stepwise-supervision split \cite{wang2024mathshepherdverifyreinforcellms} (seed 42). Each original trajectory (problem and reasoning steps) is sent unchanged to Claude-4.6-Sonnet, which independently judges the correctness of each step given the problem and preceding steps, producing a per-step true/false list. We retain an instance only when every judgment agrees with its original Math-Shepherd step label; otherwise we discard it, without rewriting any labels. This noise-filtering procedure retains 614 instances and discards 386 (38.6\%). The retained set contains 1,794 correct and 1,395 incorrect steps; 334 instances have at least one incorrect step, and 280 are all-correct. On this fixed set, SOPO averages 89.02 span-F1, compared with 85.89 for CAPO and 85.68 for GRPO; its nominal paired 95\% interval versus CAPO is above zero (Table~\ref{tab:ood_bootstrap}).

\section{Analysis of Hard Case on RAGTruth Test Set}
\label{sec:hard_case}

\begin{figure*}[h!]
	\centering
	\centering
	\includegraphics[width=\linewidth]{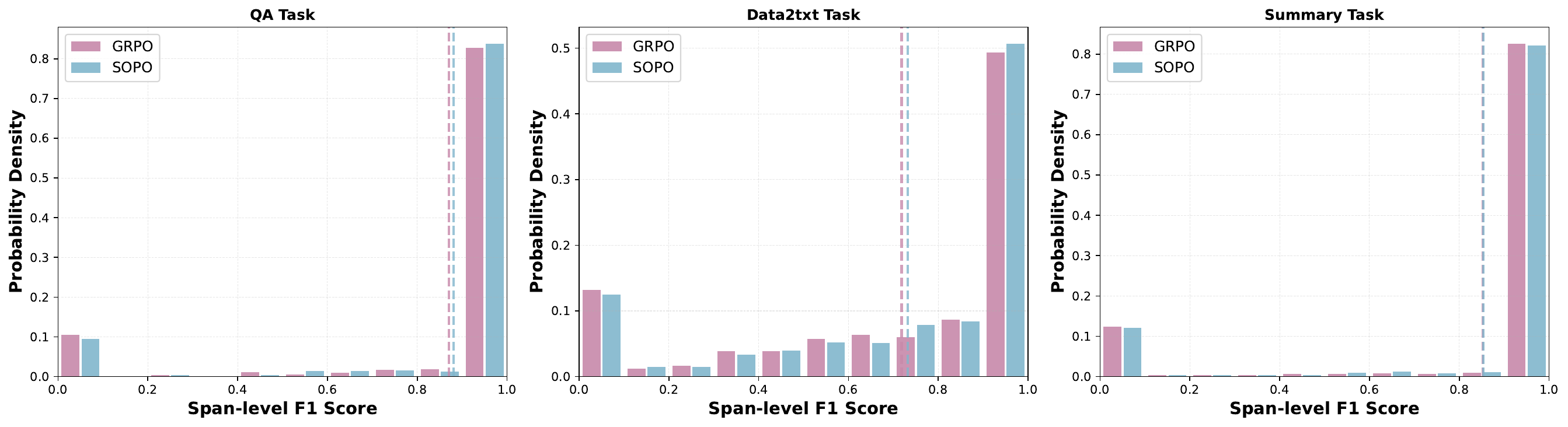}
	
	\caption{Probability density histogram of QA, data2text and summary task on RAGTruth Test set.}
	\label{fig:prob_den}
	\hfill%
\end{figure*}

\begin{figure*}[ht!]%
	\centering
	\begin{minipage}[t]{0.32\linewidth}
		\vspace{0pt}%
		\includegraphics[width=\linewidth]{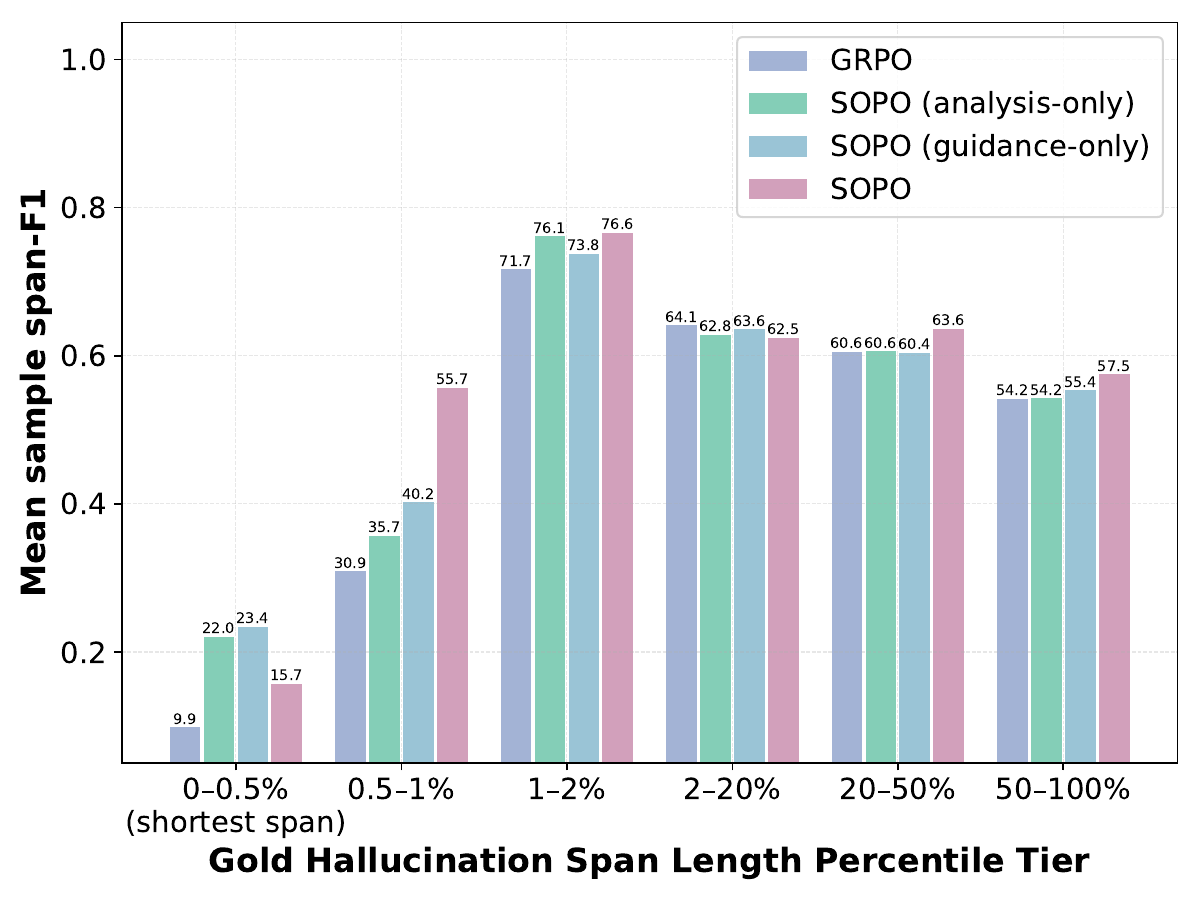}
		\caption{Mean sample span-F1 across Gold Hallucination Span Length Percentile Tiers.}
		\label{fig:hardcase_len_span}
	\end{minipage}
	\hfill
	\begin{minipage}[t]{0.32\linewidth}
		\vspace{0pt}%
		\includegraphics[width=\linewidth]{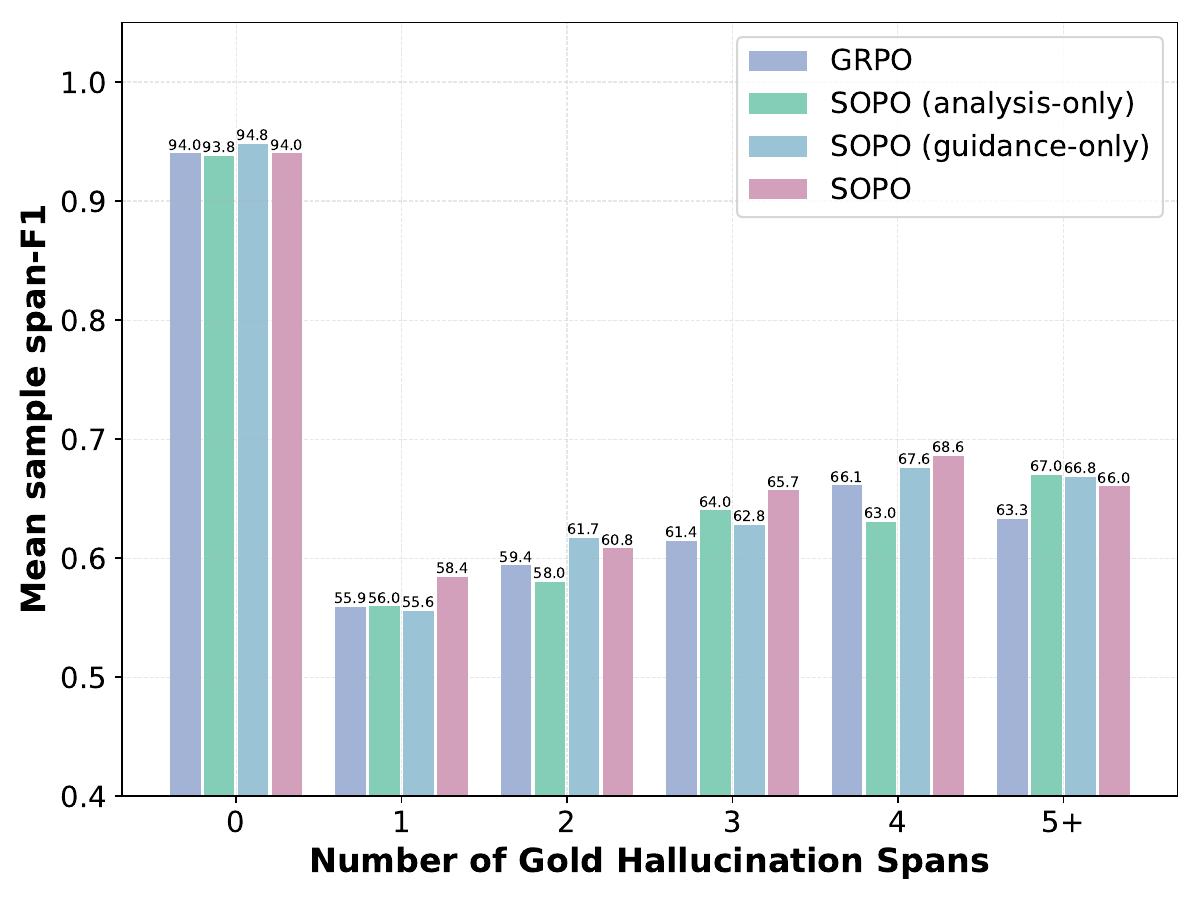}
		\caption{Mean sample span-F1 across the Number of Gold Hallucination Spans.}
		\label{fig:hardcase_n}
	\end{minipage}
	\hfill
	\begin{minipage}[t]{0.32\linewidth}
		\vspace{0pt}%
		\includegraphics[width=\linewidth]{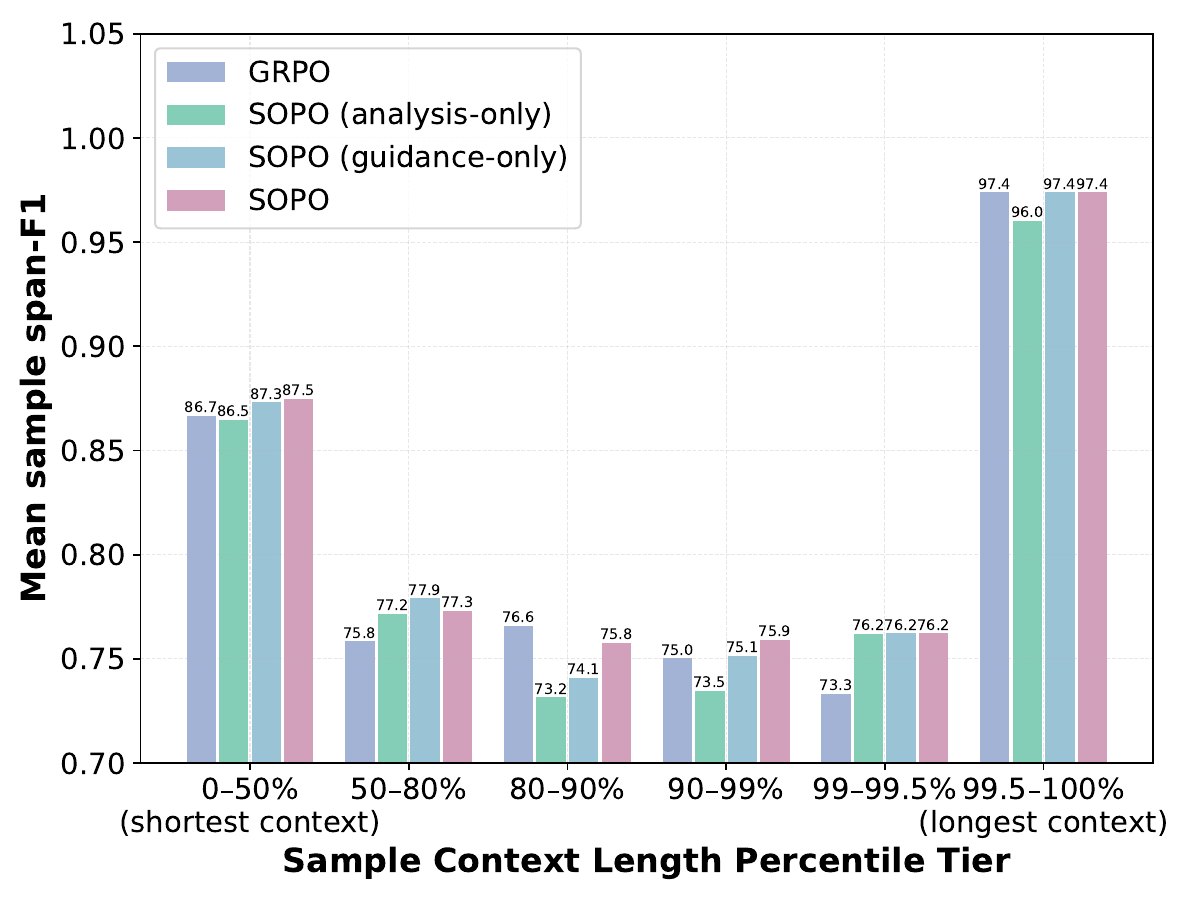}
		\caption{ Mean sample span-F1 across Sample Context Length Percentile Tiers.}%
		\label{fig:hardcase_len_context}%
	\end{minipage}
\end{figure*}

\paragraph{Breakthrough on Persistent Errors.}
We conducted a fine-grained analysis of span-level F1 scores across all 2,700 samples covering three subtasks in the RAGTruth test set. As illustrated in Figure~\ref{fig:prob_den}, SOPO exhibits a significantly lower probability density in the extremely low-score interval ($0.0-0.1$) compared to standard GRPO. This trend extends into the lower-mid range ($0.1-0.5$), highlighting a breakthrough in handling hard examples: SOPO effectively minimizes the rate of complete localization failures, converting zero-score predictions into valid detections, and driving a positive distribution shift toward higher scores. Notably, in the high-score interval ($0.8-1.0$), SOPO consistently surpasses GRPO, particularly in the QA and Data-to-Text subtasks. These results confirm that SOPO achieves superior span-level localization precision.

To provide direct evidence beyond macroscopic score histograms, we evaluated SOPO on specific ``Hard Examples'' categorized along three dimensions known to exacerbate reward sparsity and localization ambiguity:

\begin{itemize}
	\item \textbf{Samples with Short Spans (Bottom 2\% in length):} Highly sensitive cases where even slight boundary deviations drastically penalize the span-F1 score.
	\item \textbf{Samples with Multiple Spans ($\geq$3 spans):} Dense cases requiring the exhaustive detection of scattered hallucinations.
	\item \textbf{Samples with Long Contexts (Top 10\% in length):} Information-heavy cases that challenge the model's ability to filter massive amounts of noisy context.
\end{itemize}

\vspace{0.5em}
\subsection{Performance on Short Spans}

\paragraph{Numerical Errors (0--0.5\% length).} As shown in Figure \ref{fig:hardcase_len_span}, when a hallucination involves merely a mismatched number or date, the task reduces to pure string matching rather than logical deduction; consequently, the full Socratic chain offers limited marginal benefit.

\paragraph{Logical Phrases (0.5--1\% length).} For subtle logical errors, however, full SOPO proves critical. While the standard GRPO collapses to an F1 of 0.309 and the ``Guidance-only'' variant plateaus at 0.402, full SOPO achieves 0.557 (a 38.5\% relative improvement over Guidance-only). This proves that simply asking a heuristic question leaves the student guessing; explicit diagnostic reasoning (``Analysis'') is strictly required to anchor precise localization.

\subsection{Performance on dense and sparse spans}
In Figure \ref{fig:hardcase_n}, counterintuitively, our analysis reveals that the most persistent ``hard samples'' are often those containing exactly one hallucinated span, as the model struggles to suppress false positives in otherwise factual text. In this sparse setting, the ``Guidance-only'' variant (0.556 F1) actually underperforms standard GRPO (0.559 F1), demonstrating that heuristic questioning can induce over-correction without explicit diagnostic anchoring. Full SOPO (0.584 F1) successfully overcomes this limitation via its systematic ``Analysis'', achieving the highest performance in both sparse (1 span) and highly dense ($\geq$4 spans) scenarios.

\subsection{Performance on Long Context}
In Figure \ref{fig:hardcase_len_context}, while detection performance generally degrades as context length increases, the difficulty curve is not perfectly linear. Across the most challenging length percentiles (50\%--99.5\%), full SOPO consistently outperforms GRPO and all ablated variants, demonstrating robust noise-filtering capabilities and sustained attention over extended reasoning trajectories.

\section{Advantage Scaling}
\label{sec:adv_scale}
\paragraph{Pitfalls of Heuristic Post-Normalization Advantage Scaling.}
GRPO relies on group-wise normalization to ensure that the mean advantage within each group is zero. This property acts as a dynamic baseline, minimizing gradient variance. 
While recent research suggests that scaling normalized advantages is feasible under strict constraints to correct the group-relative advantage estimator bias \cite{yang2026grouprelativeadvantagebiased}, applying arbitrary heuristic scaling after normalization (such as class-aware advantage scaling from CAPO) is mathematically unsound. 

Specifically, if the scaling factor is correlated with the advantage (e.g., penalizing longer chains which might have higher/lower rewards), the weighted sum of the final advantages diverges from zero:
\begin{equation}
	\mathbb{E}[A'_{\text{i}}] \neq 0 \quad \text{where} \quad A'_{\text{i}} = \alpha_i \cdot A_{\text{norm}}
\end{equation}
This reintroduces a {baseline bias} into the gradient estimator. Unlike theoretically derived corrections that aim to restore unbiasedness, heuristic post-norm scaling arbitrarily distorts the optimization objective, leading to training instability. As illustrated in Figure~\ref{fig:reward}, SOPO with post-normalization advantage scaling exhibits severe instability around step 100, accompanied by a sharp decline in reward.

\paragraph{Rationale for Pre-Normalization Intervention.}
In contrast, our method applies the scaling factor $\alpha_i$ directly to the raw reward \textit{prior to} normalization:
\begin{equation}
	A_{\text{i}} = \text{Normalize}(\alpha_i \cdot R_{\text{i}})
\end{equation}
This distinction is crucial: we intervene at the {valuation level} rather than the estimator level. By reshaping the reward before normalization, we effectively lower the relative rank of verbose responses within the group. The subsequent normalization step then naturally handles the distribution, ensuring that the final advantages remain centered at zero with unit variance. This approach effectively incorporates our heuristic constraints without violating the mathematical integrity of the advantage estimator.

\section{Calibration of RL4HS Scores}
\label{sec:calib_score}

The discrepancy between our "Overall" scores (Table~\ref{tab:main_table}) and the "Average" scores reported in the original RL4HS \cite{su2025learningreasonhallucinationspan} paper arises from two specific deviations from the standard RAGTruth evaluation protocol:

\paragraph{Aggregation Mismatch (Macro vs. Micro-averaging).} RL4HS reported a simple macro-average across tasks. In this approach, the precision ($P_{i}$), recall ($R_{i}$), and F1 score ($F1_{i}$) are computed for each task independently, and the final score is the arithmetic mean of these individual F1 scores. Let $N$ be the total number of tasks. The macro-averaged F1 is formally defined and expanded as:
\begin{equation}
	F1_{macro} = \frac{1}{N} \sum_{i=1}^{N} F1_{i} = \frac{1}{N} \sum_{i=1}^{N} \frac{2 \cdot P_{i} \cdot R_{i}}{P_{i} + R_{i}}
\end{equation}
In contrast, the official benchmark and all standard baselines (e.g., LettuceDetect \cite{kovács2025lettucedetecthallucinationdetectionframework}, RAG-HAT \cite{song-etal-2024-rag}, Ragas \cite{es2025ragasautomatedevaluationretrieval}) strictly employ a global micro-average. This protocol aggregates the raw true positives ($TP$), false positives ($FP$), and false negatives ($FN$) across all tasks before computing the final metrics:
\begin{equation}
\begin{aligned}
	P_{micro} &= \frac{\sum_{i=1}^{N} TP_{i}}{\sum_{i=1}^{N} (TP_{i} + FP_{i})}, \quad R_{micro} = \frac{\sum_{i=1}^{N} TP_{i}}{\sum_{i=1}^{N} (TP_{i} + FN_{i})} \\
	F1_{micro} &= \frac{2 \cdot P_{micro} \cdot R_{micro}}{P_{micro} + R_{micro}}
\end{aligned}
\end{equation}
Because the subtasks in RAGTruth contain significantly different numbers of instances and hallucination densities, macro-averaging disproportionately weights smaller tasks, leading to skewed overall results.

\paragraph{Test Set Mismatch.} Furthermore, according to Appendix A.5 (Dataset Statistics) of the RL4HS paper, their QA task evaluation utilized only 875 test instances (160 + 715), deviating from the official 900 instances provided by RAGTruth.

To ensure an unbiased comparison, we standardized their results. We counted the ground-truth spans for each task in the official test set and scaled the QA subset by a factor of 875/900 to match their data distribution. Using their reported per-task metrics, we then mathematically reverse-engineered the global confusion matrix (deriving the exact $TP$, $FP$, and $FN$ counts for each task) to compute the correct micro-averaged Overall F1.

\section{Synergy of the Dual-Penalty Mechanism for Verbosity Control}
\label{sec:dual_penalty_synergy}

In multi-turn reinforcement learning, models often exploit reward hacking to maximize scores without genuinely solving the task. To cultivate an autonomous and accurate student policy, SOPO employs two complementary mechanisms to prevent different hacking strategies and strictly control reasoning verbosity.

\paragraph{Turn-Averaged Reward (Local Quality).}
If trained solely on the final-turn reward, the model could intentionally generate an empty or poor initial response merely to solicit the teacher's hints. By averaging span-level F1 scores across all turns, any early failure permanently degrades the overall score. This strict local quality constraint compels the model to provide its best reasoning attempt at every single step.

\paragraph{Advantage Scaling (Group-Relative Efficiency).}
The averaged reward alone cannot distinguish between trajectories with different interaction lengths but identical average scores. For example, consider a two-turn trajectory where the student scores 0.6 in the first turn and 0.7 in the second, resulting in a mean reward of 0.65. Without further intervention, this is indistinguishable from a more efficient single-turn trajectory that directly achieves a score of 0.65. Standard GRPO would assign these identical advantages, inadvertently encouraging interaction redundancy. Advantage Scaling directly penalizes the two-turn case by applying a scaling factor before group normalization. This intra-group efficiency constraint explicitly signals to the optimizer that shorter, highly autonomous trajectories hold a superior relative advantage.

\end{document}